\documentclass[10pt,journal,compsoc]{IEEEtran}
\ifCLASSOPTIONcompsoc
  \usepackage[nocompress]{cite}
\else
  \usepackage{cite}
\fi

\usepackage{comment} 
\usepackage[utf8]{inputenc} 
\usepackage[T1]{fontenc}    
\usepackage{hyperref}       
\usepackage{url}            
\usepackage{booktabs}       
\usepackage{balance}
\usepackage{nicefrac}       
\usepackage{microtype}      
\usepackage{subcaption}
\usepackage{graphicx}
\usepackage{amsmath,amssymb,amsfonts}
\usepackage{caption}
\usepackage{multirow}
\usepackage{xcolor}

\usepackage{amsthm}
\newtheorem{theorem}{Theorem}

\theoremstyle{definition}
\newtheorem{definition}{Definition}[section]

\ifCLASSINFOpdf
\else
\fi
\begin{document}

\title{Beyond Low-pass Filtering: Graph Convolutional Networks with Automatic Filtering}

\author{Zonghan Wu, 
        Shirui Pan, 
        Guodong Long, 
        Jing Jiang,
        and~Chengqi Zhang,~\IEEEmembership{Senior~Member,~IEEE}
\IEEEcompsocitemizethanks{\IEEEcompsocthanksitem Z. Wu, G. Long, J, Jiang, and C. Zhang are with 
Australian Artificial Intelligence Institute,  University of Technology Sydney, Australia. E-mail: zonghan.wu-3@student.uts.edu.au.
\IEEEcompsocthanksitem S. Pan is with Monash University. From Aug 2022, he will be with the School of Information and Communication Technology, Griffith University, Southport, QLD 4222, Australia.  E-mail: shiruipan@ieee.org (Corresponding Author).
}
\thanks{
S. Pan was supported in part by an ARC Future Fellowship  (FT210100097).
}
}

\markboth{Journal of \LaTeX\ Class Files,~Vol.~14, No.~8, July~2021}%
{Shell \MakeLowercase{\textit{et al.}}: Bare Demo of IEEEtran.cls for Computer Society Journals}

\IEEEtitleabstractindextext{%
\begin{abstract}
Graph convolutional networks are becoming indispensable for deep learning from graph-structured data. Most of the existing graph convolutional networks share two big shortcomings. First, they are essentially low-pass filters, thus the potentially useful middle and high frequency band of graph signals are ignored. Second, the bandwidth of existing graph convolutional filters is fixed. Parameters of a graph convolutional filter only transform the graph inputs without changing the curvature of a graph convolutional filter function. In reality, we are uncertain about whether we should retain or cut off the frequency at a certain point unless we have expert domain knowledge. In this paper, we propose Automatic Graph Convolutional Networks (AutoGCN) to capture the full spectrum of graph signals and automatically update the bandwidth of graph convolutional filters. While it is based on graph spectral theory, our AutoGCN is also localized in space and has a spatial form. Experimental results show that AutoGCN achieves significant improvement over baseline methods which only work as low-pass filters.

\end{abstract}

\begin{IEEEkeywords}
Graph convolutional networks, graph signal processing, graph neural networks.
\end{IEEEkeywords}}

\maketitle

\IEEEdisplaynontitleabstractindextext

\IEEEpeerreviewmaketitle

\IEEEraisesectionheading{\section{Introduction}\label{sec:introduction}}

\IEEEPARstart{O}{ver} the past years, convolutional neural networks and recurrent neural networks have achieved great success on grid data such as images and sequences.  Moving forward, the focus of researches gradually shifts from regular grid data to irregular non-Euclidean data \cite{bronstein2017geometric}. Being an important kind of non-Euclidean data, graphs are ubiquitous in the real world, such as cyber networks and social networks. Graph data describes the relationship, association, or interaction among different entities. To extract latent representations from data,  graph neural networks \cite{wu2020comprehensive,zhang2020deep,liu2021graph}, in particular graph convolutional networks, are developed specifically for graphs. Graph neural networks not only have leveled up benchmarks on conventional graph-related tasks \cite{gilmer2017neural, wang2017mgae, zhang2018link, simonovsky2018graphvae,pan2019learning, wang2020graph,wu2020openwgl, zhu2020graph}, but also has been proven to be helpful in solving many deep learning problems where structural dependencies exist \cite{ying2018graph,marcheggiani2017encoding,wang2018dynamic,wu2019graph,wu2020connecting}. 

Among graph neural networks, the study of graph convolutional networks is built upon the foundation of graph signal processing. With eigendecomposition of the graph Laplacian matrix,  a graph convolutional filter can be defined as a function of frequency (eigenvalue).  Through proper design of the filter function, a graph convolutional filter can be viewed from the spatial domain -  it smooths a node's inputs by aggregating information from the node's neighborhood. Due to the advantage of efficiency, generality, and flexibility, there is a trend of designing graph convolutional networks directly on spatial domains without considering their spectral properties. As a result, these spatial-designed approaches may only focus on the low frequency band of graph signals \cite{balcilar2020bridging}.  However, the middle and high frequency band of graph signals should not be ignored because they may contain useful information as well. 

\begin{figure*}
	\begin{subfigure}[b]{0.31\textwidth}   
		\centering 
		\includegraphics[width=\textwidth]{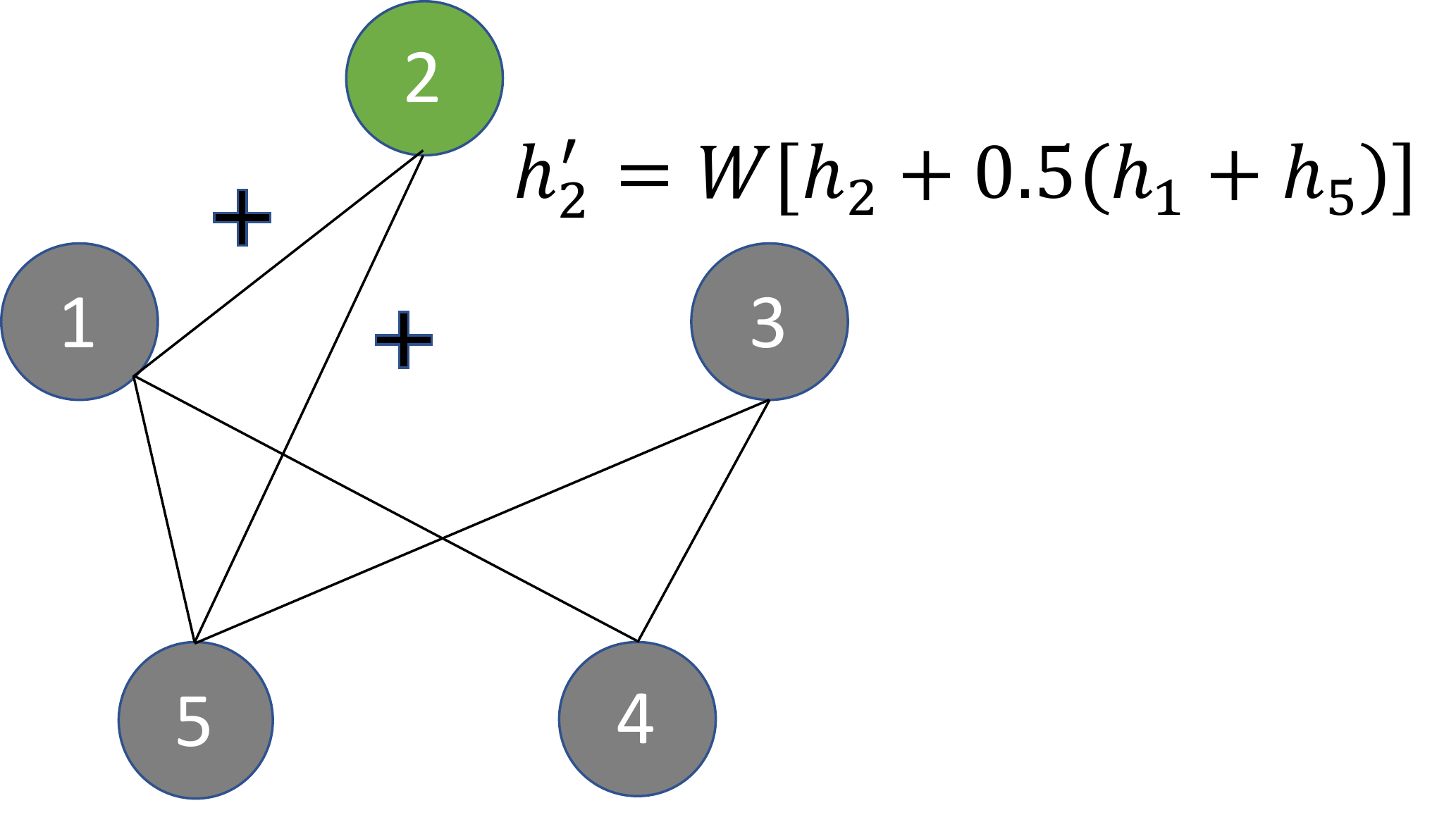}
		\caption[An example of a low-pass filter.]	
		{{\small An example of a low-pass filter. It smooths the node self information by taking the average of its neighboring information into account.}}    
		\label{fig:low}
	\end{subfigure}
	\hfill
	\begin{subfigure}[b]{0.31\textwidth}   
	\centering 
	\includegraphics[width=\textwidth]{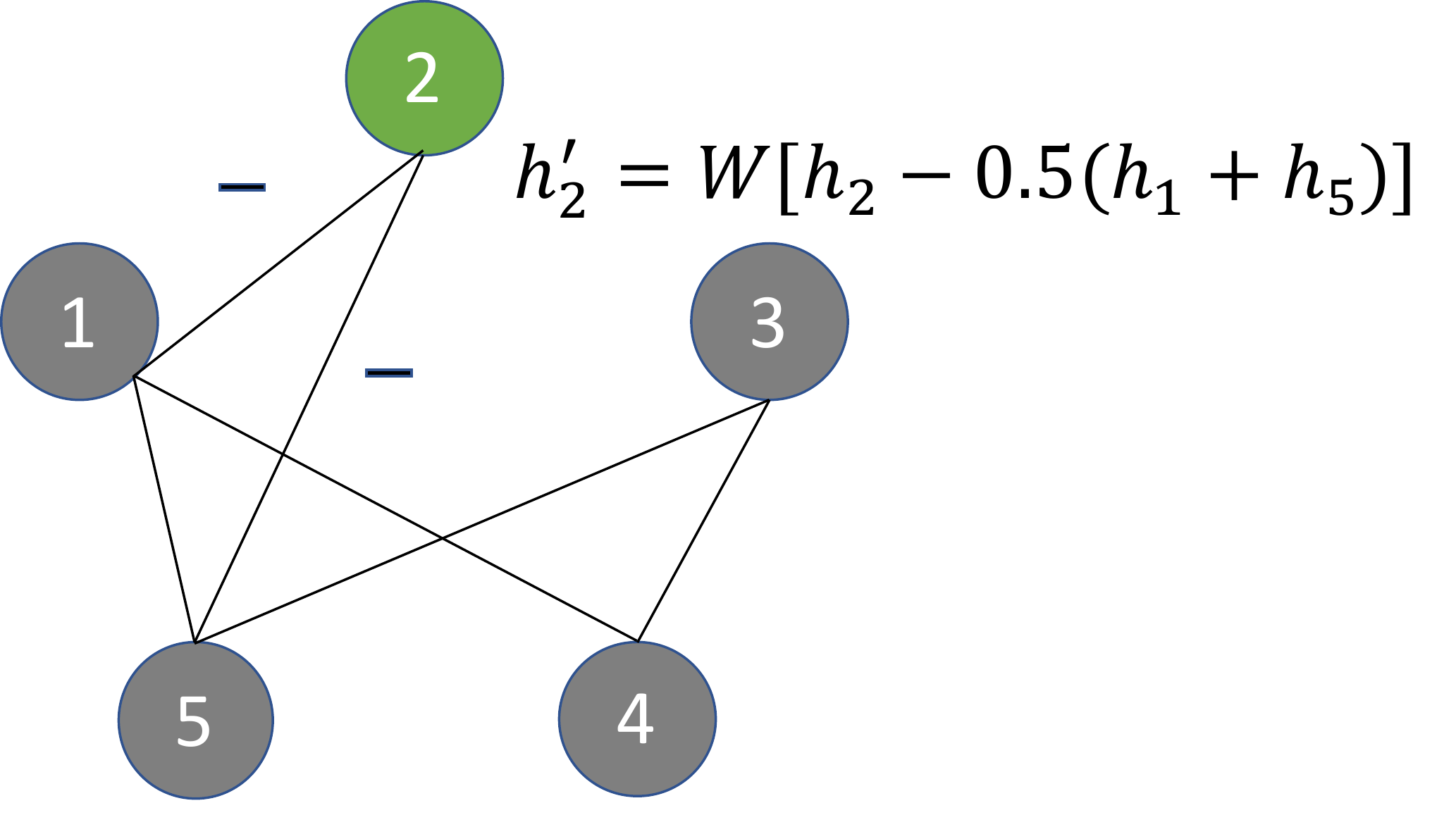}
	\caption[An example of a high-pass filter.]	
	{{\small An example of a high-pass filter. It computes the difference between the node self information with its neighboring information.}}    
	\label{fig:high}
   \end{subfigure}
   	\hfill
		\begin{subfigure}[b]{0.3\textwidth}   
		\centering 
		\includegraphics[width=\textwidth]{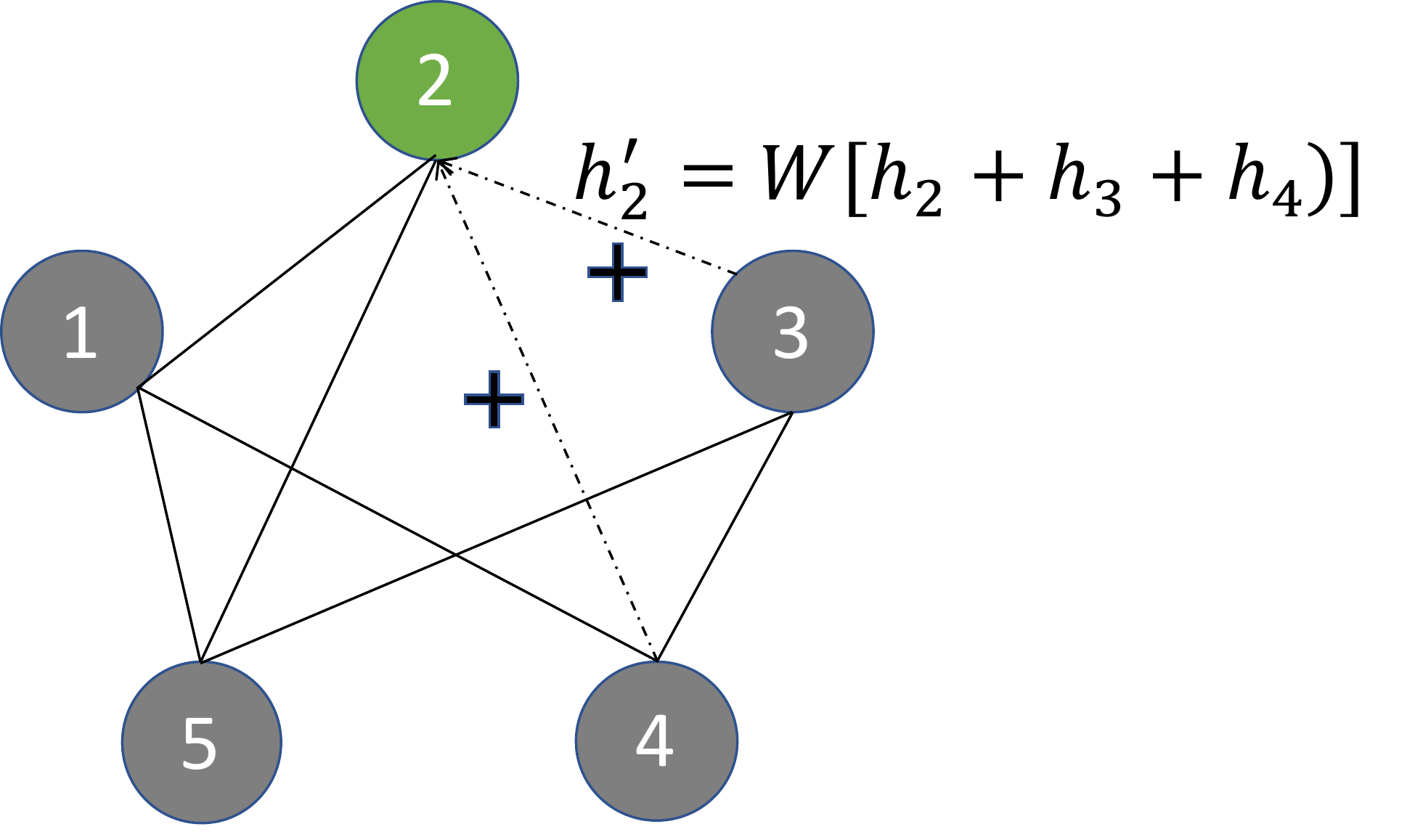}
		\caption[An example of a middle-pass filter.]
		{{\small An example of a middle-pass filter. It integrates a node self information with its second hop neighborhood information. }}    
		\label{fig:mid}
	\end{subfigure}

	\caption[Illustration of low-pass, high-pass, and middle-pass filters.]
	{\small Illustration of all-pass, low-pass, high-pass filters.} 
	\label{fig:freqpro}
\end{figure*}

To consider low, middle, high frequency band of graph signals at the same time,  we need to return back to spectral-based approaches. As a spectral-designed graph convolutional filter is defined by a function of frequency,  it can theoretically extract information on any frequency band.  Pioneer works of spectral graph convolutional networks are computationally inefficient due to eigendecomposition of the graph Laplacian matrix \cite{bruna2013spectral}.  Defferrard et al.  \cite{defferrard2016convolutional} proposed ChebNet with the filter function defined as Chebyshev polynomials. With graph Fourier transform and inverse graph Fourier transform, the graph convolutional operation essentially is reduced to multiplying linear transformed graph inputs with non-linear transformed graph Laplacian matrix. In this way,  eigendecomposition is not necessarily required.  The graph kernel of ChebNet involves computing higher orders of the graph Laplacian matrix. Kipf et al. \cite{kipf2017semi} further proposed GCN which is a first-order approximation of ChebNet with a renormalization trick. Despite that GCN improves over ChebNet in terms of efficiency,  GCN is shown to be a low-pass filter \cite{balcilar2020bridging}. 
Follow-up works of GCN also remained the same problem \cite{li2018adaptive,zhuang2018dual,wu2019simplifying}. Most recently, Balcilar et al. \cite{balcilar2020bridging} proposed Depthwise Separable Graph Convolution Network (DSGCN) with arbitrary graph convolutional filter functions, while taking low, middle, and high frequency band of graph signals into consideration.  However, filter functions of DSGCN are customized differently according to datasets without any rules to follow. It requires tremendous effort to find the optimal filter function manually for a new dataset.  More importantly, the bandwidth of graph convolutional filters of DSGCN is fixed, which also applies to other existing spectral graph convolutions. Parameters of a graph convolutional filter only transform graph inputs without changing the curvature of a graph convolutional filter function.  Therefore, the cut-off frequency or the bandwidth of a graph convolutional filter remains unchanged throughout learning.  In reality, we are uncertain about whether we should retain or cut off the frequency at a certain point unless we have expert domain knowledge. 

In this paper, we aim to design a graph convolutional network that can capture the whole spectrum of graph signals in a more efficient and effective way.  Our method consists of three graph convolutional filters, a low-pass, a middle-pass, and a high-pass filter. To avoid eigendecomposition, we limit the choice of filter functions within linear and quadratic forms of the graph Laplacian matrix. More specifically, the low-pass and high-pass filters are designed to be linear functions while the middle-pass filter has a quadratic form.  \textcolor{black}{The  roles of low-pass, high-pass, and middle-pass filter are illustrated in Figure \ref{fig:freqpro}.} Different from existing spectral-based methods, we introduce extra parameters to control the curvature and scope of all three filters. As a benefit, the bandwidth and magnitude of our graph convolutional filters can be adjusted automatically during training.

The main contributions of this paper are summarized as follows, 
\begin{itemize}
\item We propose an Automatic Graph Convolutional Network (AutoGCN) with three novel graph convolutional filters, a low-pass linear filter, a high-pass linear filter, and a middle-pass quadratic filter.  While capturing the whole spectrum of graph signals, AutoGCN ends up with a spatial form without performing eigendecomposition. 
\item We enable the proposed graph convolutional filters to control their bandwidth and magnitude automatically by updating the curvature and scope of filter functions during training. We empirically show that all three graph filters contribute to model performance.
\item Experimental results show that AutoGCN achieves significant improvement over baseline methods that only function as low-pass filters on medium-scale datasets for both node classification and graph prediction tasks. 
\end{itemize}

Our source codes are publicly available at \href{https://github.com/nnzhan/AutoGCN}{https://github.com/nnzhan/AutoGCN}.  The rest of this paper is organized as follows. In Section \ref{sec:related}, we summarize current works of graph convolutional networks. In Section \ref{sec:problem}, we formally define our problems. In Section \ref{sec:spectral}, we provide the background knowledge about spectral-rooted spatial graph convolution. In Section \ref{sec:automatic}, we present our method named automatic graph convolution in detail. In Section \ref{sec:exp}, we report the experimental results of our method on both node classification and graph prediction tasks. Finally, in Section \ref{sec:conclusion}, we make a conclusion of our paper.

\section{Related Work}
\label{sec:related}
The study of graph convolutional networks is rooted in graph signal processing or spectral graph theory \cite{shuman2013emerging}. By defining the graph Fourier transform and the inverse graph Fourier transform of a graph signal, the convolution between a graph signal and a filter can be derived by the convolution theorem where the Fourier transform of the convolution of two signals equals the elementwise product of their Fourier transforms. With this theorem,  Bruna et al. \cite{bruna2013spectral} defined graph convolutional filters as functions of eigenvalues of the graph Laplacian matrix. Eigendecomposition of a graph Laplacian matrix is computationally expensive. To bypass this bottleneck, Defferrard et al. \cite{defferrard2016convolutional} showed that a filter function defined on the eigenvalues is equivalent to the same function defined on the graph Laplacian matrix. Based on this result, various of filter functions which are defined on the graph Laplacian matrix directly have been proposed such as Chebyshev polynomials \cite{defferrard2016convolutional}, a first-order approximation of Chebyshev polynomials \cite{kipf2017semi}, and Cayley polynomials \cite{levie2018cayleynets}. Besides graph Fourier transform, another line of works defines graph convolution through graph wavelet transform \cite{hammond2011wavelets,xu2019graph}. These methods are localized in the vertex domain and do not require eigendecomposition as well. However, complex operations on the graph Laplacian matrix still impede model efficiency due to higher-order computation. 

Concurrently, despite graph convolution, message passing has mostly dominated the recent development of graph neural networks due to its efficiency, generality, and flexibility. The basic idea is to propagate graph signals along with graph structures. By iterating the propagation step multiple times, a node can broaden its neighborhood to the entire graph. Earlier schemes of graph message passing follow the recurrent architecture, where the parameters are shared across multiple propagation steps \cite{scarselli2009graph,li2015gated,dai2018learning}. These methods update node states recursively until steady states are reached. On the other side, compositional schemes modularize message passing as a neural network layer to improve model pluggability and capacity \cite{micheli2009neural}. Gilmer et al. \cite{gilmer2017neural} formalized the message passing framework with two components, a propagator and a updator. The propagator summarizes information for a node based on its neighborhood context. The updator transforms the collected information with learnable parameters.  The most intuitive propagate function is the mean function, which takes the average of the center node information and its neighborhood information. Various improvements are made over the mean aggregate function by refining edge weights through normalization \cite{kipf2017semi} and diffusion \cite{atwood2016diffusion}, assigning learnable weights to neighbors \cite{monti2017geometric,velickovic2017graph}, introducing a scalar parameter to pass graph isomorphism test \cite{xu2018how}, teleporting back to a node's initial information to avoid the over-smoothing problem  \cite{klicpera2018predict}.  \textcolor{black}{In addition, the expressive power of most GNNs are upbounded by the 1-Weisfeiler-Lehman (1-WL) graph isomorphism test. Some recent works are able to surpass this limitation \cite{you2021identity, zeng2021decoupling}.  The over-smoothing problem is studied in depth by \cite{zeng2021decoupling} and \cite{chen2020simple}. They introduced decoupling mechanisms and identity mapping respectively.} Some other advanced works focus on designing complex architectures of graph neural networks by increasing network depth and receptive field size \cite{gao2019graph,luan2019break,abu2019mixhop}. Graph neural networks based on message passing do not consider spectral properties of the message passing operations, namely graph convolution. Wu et al. \cite{wu2019simplifying} proved that stacking multiple GCN with identity activation is a low-pass filter. Balcilar et al. \cite{balcilar2020bridging} showed that most graph neural networks are essentially low-pass filters.  \textcolor{black}{
In parallel to graph convolution transforms, graph wavelet transforms decompose graph signals in multi-scale \cite{coifman2006diffusion,gao2019geometric}. These works also consider the high-frequency part of graph signals, but they lack the ability to automatically adjust the bandwidth of graph signals based on data. }
\section{Problem Formulation}
\label{sec:problem}

\textbf{Attributed graph.} An attributed graph $G=(V,E,\mathbf{X})$ consists of a set of nodes $V$,  a set of edges $E$, and a node feature matrix $\mathbf{X}\in\mathbf{R}^{n\times d}$ where $n$ is the number of nodes for the graph $G$  and $d$ is the feature dimension.  The structural information of the graph $G$ can be encoded by a graph adjacency matrix $\mathbf{A}\in \mathbf{R}^{n\times n}$. If there exists a connection $(v_i,v_j)\in E$, then $\mathbf{A}_{ij}\ne 0$, otherwise $\mathbf{A}_{ij} =0$. In this paper, we only consider undirected attributed graphs. In this case, $\mathbf{A}$ is symmetric.

\textbf{Node-level prediction.} The node-level prediction task forecasts a label for each node in a graph. It aims to learn a mapping $f$:$(\mathbf{A},\mathbf{X})\rightarrow\mathbf{Y}\in \mathbf{R}^{n\times c}$. For node regression problems, $c=1$. For node classification problems, $c$ equals the number of classes.

\textbf{Graph-level prediction.} The graph-level prediction task forecasts a label for each graph in a data set. It aims to learn a mapping $f$:$(\mathbf{A},\mathbf{X})\rightarrow\mathbf{Y}\in \mathbf{R}^{1\times c}$. For graph regression problems, $c=1$. For graph classification problems, $c$ equals the number of classes.

\section{Spectral-Rooted Spatial Graph Convolution}
\label{sec:spectral}

 Graph convolutional networks generally fall into spectral-based approaches and spatial-based approaches.  Several representative methods such as ChebNet \cite{defferrard2016convolutional} and GCN \cite{kipf2017semi} take roots in the spectral domain while ending up with spatial forms. As a key benefit, they avoid the computation of eigendecomposition of the graph Laplacian matrix in order to perform graph convolution. Most recently, Balcilar et al. \cite{balcilar2020bridging} connect spectral-based approaches and spatial-based approaches by a uniform formula, which is defined as
\begin{equation}
\label{eq:gcn1}
\mathbf{H}^{(l+1)}=\sigma(\sum_{k=1}^{K} \mathbf{C}^{(k)}\mathbf{H}^{(l)}\mathbf{W}^{(l,k)})
\end{equation}
with the graph convolutional kernel set to 
\begin{equation}
\label{eq:gcn2}
	\mathbf{C}^{(k)} = \mathbf{U}diag(F_k(\boldsymbol\lambda))\mathbf{U}^T
\end{equation}
where $F_k(\cdot)$ is the filter function,  $\mathbf{U}$ and $\boldsymbol\lambda$ denote the eigenvectors and  the eigenvalues of the normalized graph Laplacian matrix $\mathbf{L}=\mathbf{I}-\mathbf{D}^{-\frac{1}{2}}\mathbf{A}\mathbf{D}^{-\frac{1}{2}}$ respectively, $\mathbf{D}$ is a diagnal matrix with $\mathbf{D}_{ii}=\sum_j \mathbf{A}_{ij}$,  $\mathbf{W}^{(l,k)}$ is a learnable parameter matrix, $\mathbf{H}^{(l)}$ is the node representation matrix at layer $l$, $\mathbf{H}^{(1)}=\mathbf{X}$,  $K$ is the number of filter functions, and $\sigma(\cdot)$ represents the activation function.  With Equation \ref{eq:gcn1} and \ref{eq:gcn2}, spectral-based apporaches can be designed by defining new filter functions, and spatial-based approaches can be analyzed from the spectral domain by computing the frequency profile (the filter function) of the convolutional kernel $\mathbf{C}^{(k)}$,  using 
\begin{equation}
	F_k(\boldsymbol\lambda)= diag(\mathbf{U}^T\mathbf{C}^{(k)}\mathbf{U}).
\end{equation}

Under this framework, ChebNet takes the filter function $F_1(\boldsymbol\lambda)=\mathbf{1}$, $F_2(\boldsymbol\lambda)=2\boldsymbol\lambda/\lambda_{max}-\mathbf{1}$, and $F_k(\boldsymbol\lambda)= 2F_2(\boldsymbol\lambda)F_{k-1}(\boldsymbol\lambda)-F_{k-2}(\boldsymbol\lambda)$.  As $\mathbf{L}=\mathbf{U}diag(\boldsymbol\lambda)\mathbf{U}^T$, the convolutional kernel of ChebNet, i.e. $\mathbf{C}^{(k)}$,  results in a polynomial function of $\mathbf{L}$ with order $k-1$. Taking higher orders of $\mathbf{L}$ is computationally expensive. GCN simplifies ChebNet with first order approximation by setting $K=2$,  $\lambda_{max}=2$, and $\mathbf{W}^{(l,1)}=-\mathbf{W}^{(l,2)}$.  The form of GCN is then derived as, 
\begin{align}
	\mathbf{H}^{(l+1)} &= \sigma(\mathbf{I}\mathbf{H}^{(l)}\mathbf{W}^{(l,1)}-(\mathbf{L}-\mathbf{I})\mathbf{H}^{(l)}\mathbf{W}^{(l,1)}) \\
	&=\sigma((\mathbf{I}+\mathbf{D}^{-\frac{1}{2}}\mathbf{A}\mathbf{D}^{-\frac{1}{2}})\mathbf{H}^{(l)}\mathbf{W}^{(l,1)}) 
\end{align}
With a renormalization trick to avoid numerical instabilities, GCN replaces $\mathbf{I}+\mathbf{D}^{-\frac{1}{2}}\mathbf{A}\mathbf{D}^{-\frac{1}{2}}$ by $\mathbf{\tilde{D}}^{-\frac{1}{2}}(\mathbf{A}+\mathbf{I})\mathbf{\tilde{D}}^{-\frac{1}{2}}$, where $\mathbf{\tilde{D}}_{ii}=\sum_j\mathbf{A}_{ij}+1$. Proved by Balcilar et al. \cite{balcilar2020bridging}, the frequency profile of GCN can be approximated by $F(\boldsymbol\lambda)=\mathbf{1}-\boldsymbol\lambda\bar{d}/(1+\bar{d})$. This corresponds to a low-pass filter with the cut-off frequency $(1+\bar{d})/\bar{d}$. Although GCN is simple and robust, it only contains a single low-pass filter. It is incapable of capturing potential patterns in middle and high frequency bands.  DSGCN \cite{balcilar2020bridging} incorporates low-pass, middle-pass and high-pass filters by proposing a set of customized filter functions $F(\boldsymbol\lambda)$ for a particular dataset. For example, DSGCN proposes four filter functions for the CORA dataset. They are $F_1(\boldsymbol\lambda)=(1-\boldsymbol\lambda/\lambda_{max})^5$, $F_2(\boldsymbol\lambda)=exp(-0.25(0.25\lambda_{max}-\boldsymbol\lambda)^2)$,  $F_3(\boldsymbol\lambda)=exp(-0.25(0.5\lambda_{max}-\boldsymbol\lambda)^2)$, and
$F_4(\boldsymbol\lambda)=exp(-0.25(0.75\lambda_{max}-\boldsymbol\lambda)^2)$.  While capturing customized frequency band of graph signals, DSGCN needs to calculate eigenvalues and eigenvectors, causing scalability issues. Furthermore, the filter functions of DSGCN are manually designed. It lacks an automatic mechanism to learn the optimal frequency profile of filters based on data.

\section{Automatic Graph Convolution}
\label{sec:automatic}

Based on spectral-rooted spatial graph convolution, we propose Automatic Graph Convolution that automatically selects low-pass, middle-pass, and high-pass filters with optimal frequency profiles.  To avoid heavy computations of eigenvalues or higher orders of the graph Laplacian matrix $\mathbf{L}$,  we restrict the search of filter functions within linear and quadratic forms. 

In the following, we first introduce the proposed low-pass, high-pass, and middle-pass filters, and then elaborate on the design of our proposed graph convolution. 

Our low-pass and high-pass filters are designed to be a linear function of $\boldsymbol\lambda$. We introduce two adjustable parameters to control the magnitude and the cut-off frequency of a filter function.  We propose the \textbf{low-pass linear filter} function  as 
\begin{equation}
\label{eq:low}
	F_{low}(\boldsymbol\lambda) = p(1-a\boldsymbol\lambda),
\end{equation}
where the parameter $p>0$ controls the magnitude of the frequency profile and  the parameter $a\in (0,1)$ determines the cut-off frequency.  \textcolor{black}{The reason we constrain the scope of the parameter $a$ within $a\in(0,1)$ is based on the assumption that  $\lambda_{max}=2$, as the upper bound of eigenvalues is $2$.} When $a$ approaches to $1$, at least half of the lower spectrum can be retained. If $a$ approaches to $0$, the slope of the low-pass filter function will become flatter in a decreasing manner.  In Figure \ref{fig:low}, we give  examples of the frequency profiles of low-pass linear filters with three different settings.  Inserting Equation \ref{eq:low} into Equation \ref{eq:gcn2}, the graph convolutional kernel of the low-pass linear filter is derived as 
\begin{equation}
\label{eq:clow}
		\mathbf{C}_{low}(p,a) = p(a\tilde{\mathbf{A}}+(1-a)\mathbf{I})
\end{equation}

where $\tilde{\mathbf{A}}=\mathbf{D}^{-\frac{1}{2}}\mathbf{A}\mathbf{D}^{-\frac{1}{2}}$. 
The derivation process can be found in Appendix A. According to Equation \ref{eq:clow}, the low-pass linear filter essentially aggregates a node's self-information with its neighborhood information. It is  reasonable for graph data which follows the homophily assumption that connected nodes share similar features. From a spatial perspective, the parameter $p$ controls the contribution weight of a low-pass filter, and $a$ adjusts the confidence level of the homophily assumption. 

\begin{figure*}
	\begin{subfigure}[b]{0.3\textwidth}   
		\centering 
		\includegraphics[width=\textwidth]{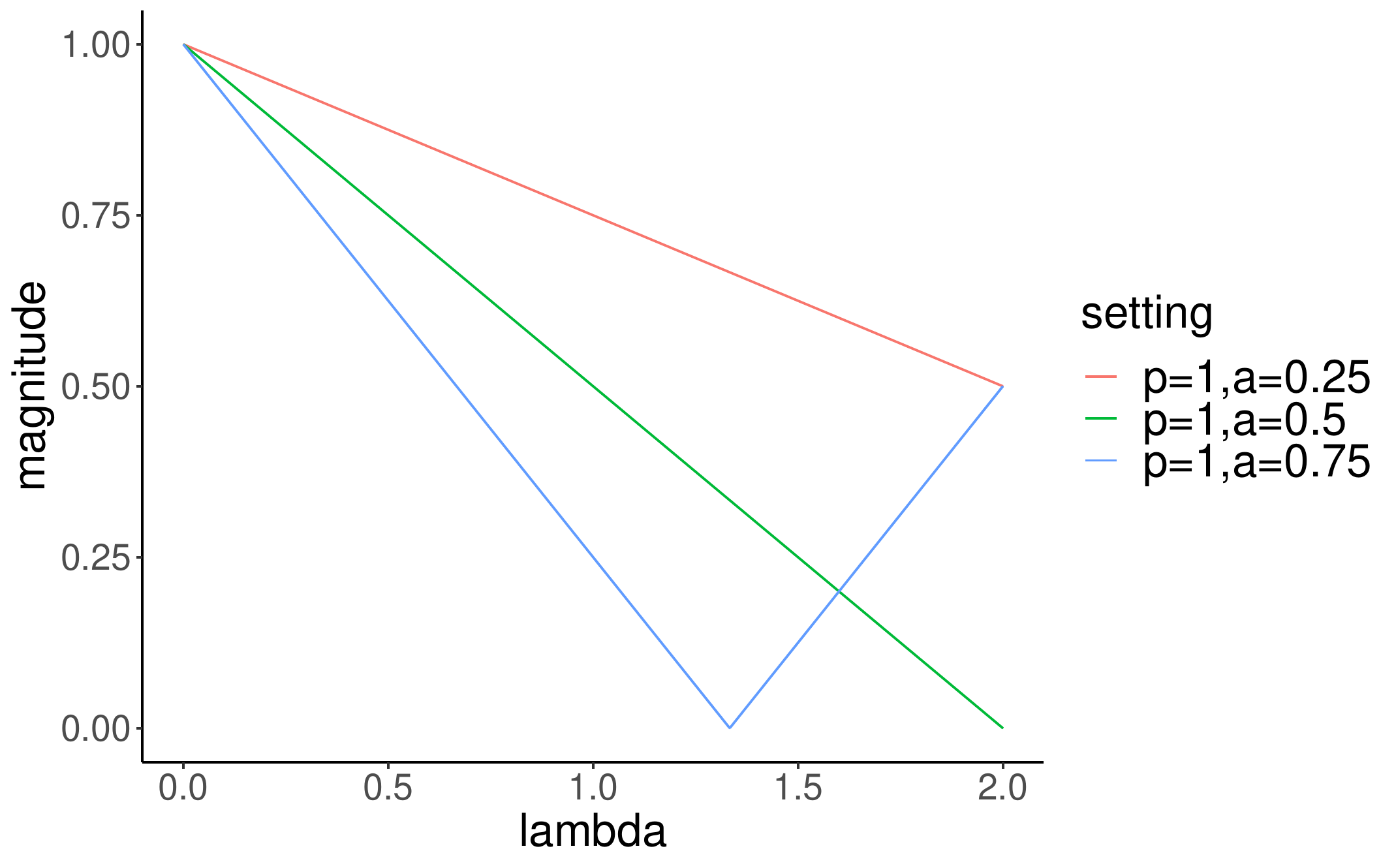}
		\caption[The frequency profiles of the low-pass filters.]
		{{\small The frequency profiles of the low-pass filters.}}    
		\label{fig:low}
	\end{subfigure}
	\hfill
	\begin{subfigure}[b]{0.3\textwidth}   
		\centering 
		\includegraphics[width=\textwidth]{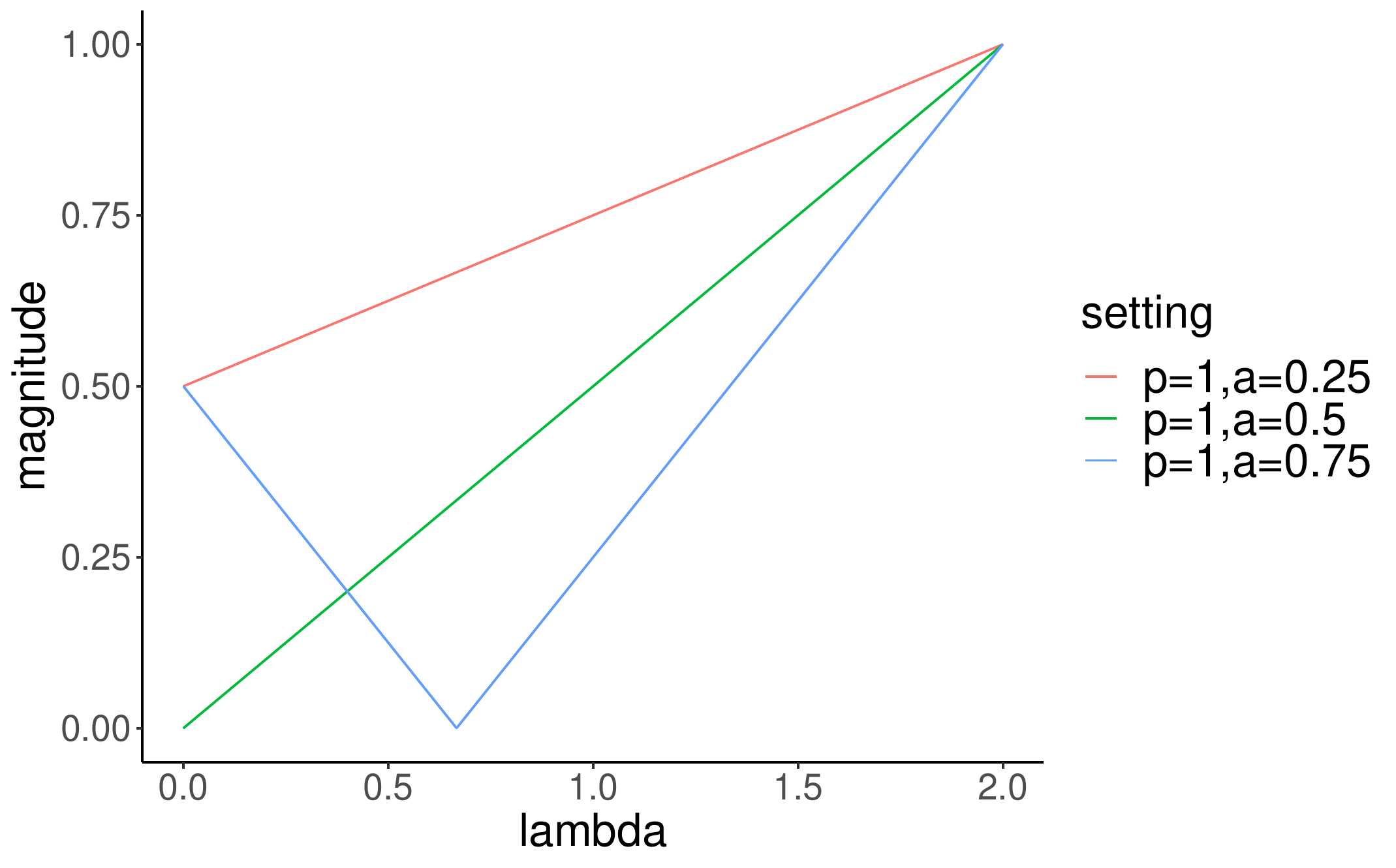}
		\caption[The frequency profiles of the high-pass filters.]	
		{{\small The frequency profiles of the high-pass filters.}}    
		\label{fig:high}
	\end{subfigure}
	\hfill
	\begin{subfigure}[b]{0.3\textwidth}   
	\centering 
	\includegraphics[width=\textwidth]{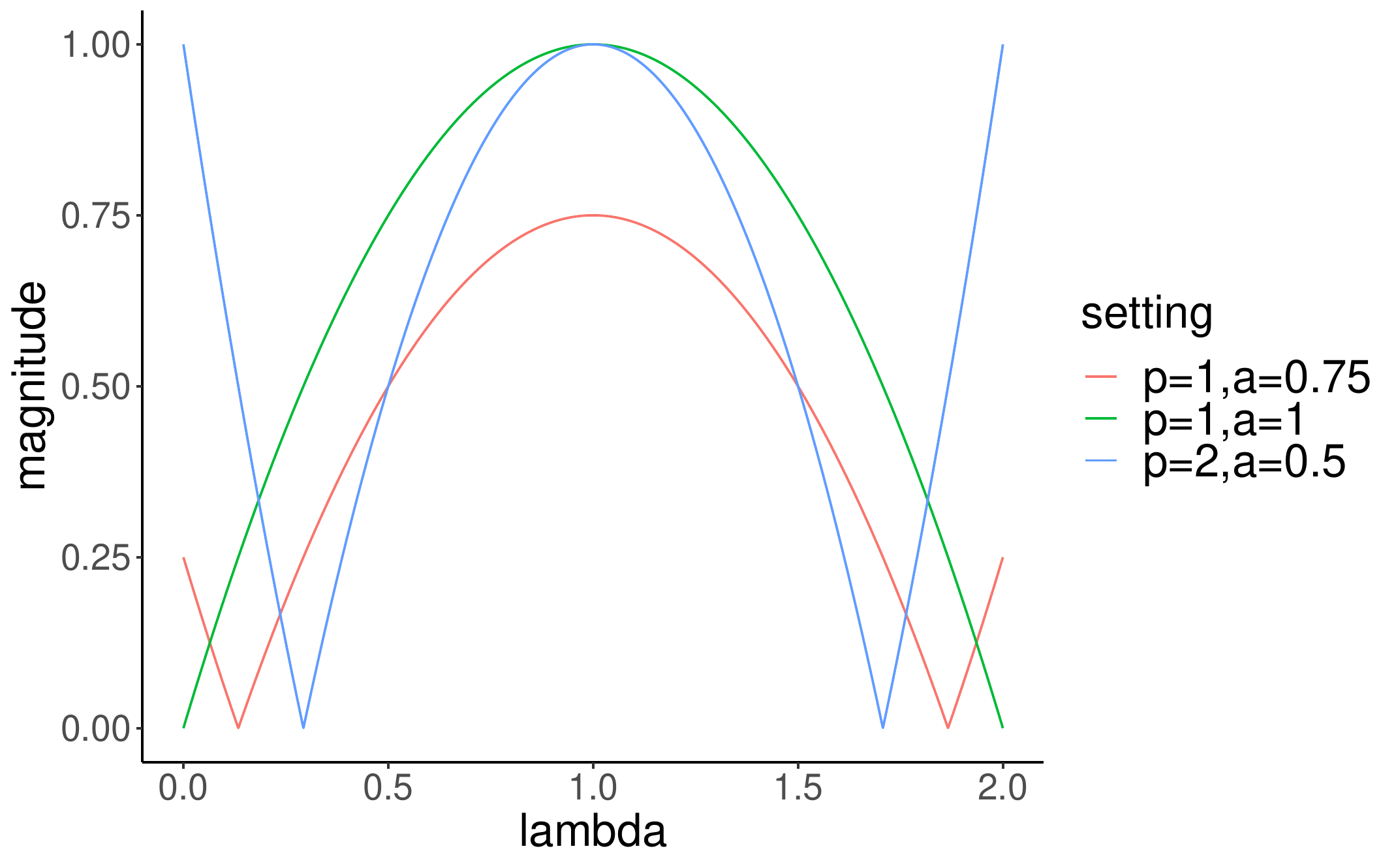}
	\caption[The frequency profiles of the middle-pass filters.]	
	{{\small The frequency profiles of the middle-pass filters.}}    
	\label{fig:mid}
   \end{subfigure}

	\caption[Frequency profiles of low-pass, high-pass, and middle-pass filters with three different settings.]
	{\small Frequency profiles of low-pass, high-pass, and middle-pass filters with three different settings. The horizontal axis is $\lambda$, ranging from $0$ to $2$. The vertical axis is the magnitude, which is the absolute value of the filter function.} 
	\label{fig:freqpro}
\end{figure*}

Different from a low-pass filter,  the spectrum of a high-pass filter should be retained in a increasing manner.  We propose the \textbf{high-pass linear filter} function  as
\begin{equation}
\label{eq:high}
	F_{high}(\boldsymbol\lambda) = p(a\boldsymbol\lambda+1-2a),
\end{equation}
where $p>0$ and $a\in (0,1)$.  We can easily derive that when $\lambda=2$, the filter function achieves the highest magnitude $p$ and when $\lambda$ decreases to $2-1/a$, it reaches $0$. In Figure \ref{fig:high}, we give examples of the frequency profiles of high-pass linear filters with three different settings. Placing Equation \ref{eq:high} into Equation \ref{eq:gcn2}, the graph convolutional kernel of the proposed high-pass linear filter is derived as 
\begin{equation}
\label{eq:chigh}
\mathbf{C}_{high}(p,a) = p(-a\tilde{\mathbf{A}}+(1-a)\mathbf{I}).
\end{equation}
We provide the details of derivation in Appendix A.  From Equation \ref{eq:chigh},  the high-pass filter computes the difference between the self-information and neighborhood information. It highlights the features of a node that are distinct from its neighbors. 

The middle-pass filter cuts off frequency values at the low and high end.  Due to linear functions are incapable of capturing this property, we assume the middle-pass filter function has a quadratic form.
We propose the \textbf{middle-pass quadratic} filter function as 
\begin{equation}
\label{eq:middle}
	F_{mid}(\boldsymbol\lambda) = p((\boldsymbol\lambda-1)^2-a),
\end{equation}	
where $p>0$ and $a\in(0,1]$. The proposed middle-pass filter cuts off frequency at $\lambda=1 \pm \sqrt{a}$, and reaches its maximum at $\lambda= 1$.  Figure \ref{fig:mid} shows examples of the frequency profiles of the middle-pass filters with three different settings.  Taking Equation \ref{eq:middle} into Equation \ref{eq:gcn2},  the graph convolutional kernel of the proposed middle-pass filter is derived as, 

\begin{equation}
\label{eq:cmid}
\mathbf{C}_{mid}(p,a) = p(\tilde{\mathbf{A}}^2-a\mathbf{I}).
\end{equation}

Viewing from a spatial perspective,  our middle-pass filter can be interpreted as differentiating  a node's self-information from its two-hop neighborhood information.

\begin{theorem} 
Assume a set of base low-pass linear filter functions with $F_i(\boldsymbol\lambda)=1-a_i\boldsymbol\lambda$  $(i=1,2,\cdots,K)$, $a_i\in (0,1)$,  the linear combination of the set of base low-pass linear filter functions, $F(\boldsymbol\lambda) =\sum p_i  F_i(\boldsymbol\lambda)$, with $p_i>0$, is a low-pass linear filter with $\tilde{p}=\sum p_i$ and $\tilde{a}=\sum p_ia_i/\sum p_i$.
\end{theorem}

\begin{proof}
\begin{align}
F(\boldsymbol\lambda) &=\sum p_i  F_i(\boldsymbol\lambda) \\
& = \sum p_i(1-a_i\boldsymbol\lambda) \\
& = \sum p_i \times (1-\frac{\sum p_ia_i}{\sum p_i}\boldsymbol\lambda)
\end{align}
Let $\tilde{p}=\sum p_i$, and $\tilde{a}=\sum p_ia_i/\sum p_i$, 
\begin{align}
\label{eq:lowp}
F(\boldsymbol\lambda) &=\tilde{p}(1-\tilde{a}\boldsymbol\lambda )
\end{align}
We now prove $\tilde{p}>0$, and $\tilde{a} \in (0,1)$.
\begin{align}
\tilde{p} &=\sum p_i > \min(p_i) > 0
\end{align}
As $p_i>0$,
\begin{align}
\tilde{a} = \sum p_ia_i/\sum p_i \ge \sum p_i\min(a_i)/\sum p_i = min(a_i)>0
\end{align}
\begin{align}
\tilde{a} = \sum p_ia_i/\sum p_i \le \sum p_i\max(a_i)/\sum p_i = max(a_i)<1
\end{align}
Therefore, 
$0<\tilde{a}<1$.

With $p_i>0$ and $0<\tilde{a}<1$, Equation \ref{eq:lowp} fulfills the defintion of our proposed low-pass linear filter.

\end{proof}

\begin{theorem} 
 Assume a set of base high-pass linear filter functions with $F_i(\boldsymbol\lambda)=a_i\boldsymbol\lambda+1-2a_i$ $(i=1,2,\cdots,K)$,  $a_i\in (0, 1)$, the linear combination of the set of base high-pass linear filter functions, $F(\boldsymbol\lambda) = \sum_i p_i  F_i(\boldsymbol\lambda) $, with $p_i>0$,  is  a high-pass linear filter with $\tilde{p}=\sum p_i$ and $\tilde{a}=\sum p_ia_i/\sum p_i$.
\end{theorem}

\begin{proof}
\begin{align}
F(\boldsymbol\lambda) &=\sum p_i  F_i(\boldsymbol\lambda) \\
& = \sum p_i(a_i\boldsymbol\lambda+1-2a_i) \\
& = (\sum p_i) \times (\frac{\sum p_ia_i}{\sum p_i}\boldsymbol\lambda+ 1- 2\frac{\sum p_ia_i}{\sum p_i})
\end{align}
Let $\tilde{p}=\sum p_i$, and $\tilde{a}=\sum p_ia_i/\sum p_i$, 
\begin{align}
\label{eq:highp}
F(\boldsymbol\lambda) &=\tilde{p}(\tilde{a}\boldsymbol\lambda+1-2\tilde{a} )
\end{align}
As $\tilde{p}>0$ and $\tilde{a} \in (0,1)$,  Equation \ref{eq:highp} fulfills the defintion of our proposed high-pass filter.
\end{proof}

\begin{theorem} 
Assume a set of base middle-pass quadratic filter functions with $F_i(\boldsymbol\lambda)=(\boldsymbol\lambda-1)^2-a_i$ $(i=1,2,\cdots,K)$,  $a_i\in (0, 1]$, the linear combination of the set of base middle-pass quadratic filter functions, $F(\boldsymbol\lambda) = \sum_i p_i  F_i(\boldsymbol\lambda)$, with $p_i>0$,  is  a middle-pass quadratic filter with $\tilde{p}=\sum p_i$ and $\tilde{a}=\sum p_ia_i/\sum p_i$.
\end{theorem}

\begin{proof}
\begin{align}
F(\boldsymbol\lambda) &=\sum p_i  F_i(\boldsymbol\lambda) \\
& = \sum p_i((\boldsymbol\lambda-1)^2-a_i) \\
& = (\sum p_i)  \times ((\boldsymbol\lambda-1)^2- \frac{\sum p_i a_i}{\sum p_i} )\\
\end{align}
Let $\tilde{p}=\sum p_i$, and $\tilde{a}=\sum p_ia_i/\sum p_i$, 

\begin{align}
\label{eq:middlep}
F(\boldsymbol\lambda) &=\tilde{p}((\boldsymbol\lambda-1)^2-\tilde{a} )
\end{align}
As $\tilde{p}>0$ and $\tilde{a} \in (0,1]$,  Equation \ref{eq:middlep} fulfills the defintion of our proposed middle-pass filter.
\end{proof}

\textbf{Over-parameterization of low-pass, high-pass, and middle-pass filter functions.} For a single low-pass, high-pass, or a middle-pass filter, to learn the scalar parameter $p$ and $a$ by gradient descent is not robust. As $p$ and $a$ are both one-dimensional, the distance between two local optimums for these two parameters is quite small compared to points in higher-dimensional space.  Therefore $p$ and $a$ can be easily shifted from one local optimum point to another in each learning iteration. 
To address this issue,  we over-parameterize the proposed filter functions by a linear combination of base filter functions.  According to Theorem 1, 2, and 3,  the linear combination of base filter functions still lies in the definition of our proposed filter functions.  Concretely, we set the over-parameterized low-pass, high-pass, and middle-pass filter functions as $F_{low}(\boldsymbol\lambda)= \sum_{i=1}^{K}p_i(1-a_i\boldsymbol\lambda)$, $F_{high}(\boldsymbol\lambda)=\sum_{i=1}^{K}p_i(a_i\boldsymbol\lambda+1-2a_i)$, and $F_{mid}(\boldsymbol\lambda) = \sum_{i=1}^{K}p_i((\boldsymbol\lambda-1)^2-a_i)$, where $K$ denotes the number of base filter functions, $p_i$ is a learnable parameter constrained by $p_i>0$, $\{a_i, i=1,2,\cdots,K\}$ are set to a fixed value, which equally spaced across $(0,1)$ for low-pass and high-pass filter functions, and across $(0,1]$ for middle-pass filter functions.  For example, if $K=3$ and $a_i\in(0,1)$, then $\{a_1,a_2,a_3\}$ is set to $\{0+\epsilon,0.5-\epsilon,1-\epsilon\}$ with $\epsilon$ set to an infinitely small value.  By setting a large $K$, we essentially over-parameterize and transform the proposed filter functions from learning two single parameters ($p$ and $a$)  to learning a set of parameters ($\{p_i, i=1,2,\cdots,K\}$).  In this way,  the model generalization power is enhanced. 

\textbf{Complementary gating.} Gating mechanisms are widely used in neural networks to control information flow and increase non-linearity. Without additional parameters, we introduce a complementary gating mechanism. The idea is to weigh the importance of one graph convolutional filter given the other twos. We assume that if one of the three graph convolutional filters contributes to the learning objective, the role of the other twos will be less important.
Combining all components,  we reach to our proposed graph convolution named \textbf{Automatic Graph Convolution} in its spatial form:  

\begin{align*}
    \mathbf{H}^{(l+1)}&=\sigma(\mathbf{H}^{(l)}_{low}\odot\sigma(\mathbf{H}^{(l)}_{high}+\mathbf{H}^{(l)}_{mid})+ \\
& \mathbf{H}^{(l)}_{high}\odot\sigma(\mathbf{H}^{(l)}_{low}+\mathbf{H}^{(l)}_{mid})+\mathbf{H}^{(l)}_{mid}\odot\sigma(\mathbf{H}^{(l)}_{low}+\mathbf{H}^{(l)}_{high}))
\end{align*}

where $\odot$ denotes the elementwise product, $\mathbf{H}^{(l)}_{f}=\mathbf{C}^{(l)}_{f}\mathbf{H}^{(l)}\mathbf{W}^{(l)}_{f}$, $\mathbf{C}^{(l)}_{f}=\mathbf{U}diag(F_f(\boldsymbol\lambda))\mathbf{U}^T$, and $f\in\{low,mid,high\}$. To form a graph convolutional network, the automatic graph convolution can be performed multiple times attached with an output layer in the end.  For node prediction tasks, the output layer can be an MLP (multi-layer perceptron). For graph prediction tasks, the output layer can be a sum/mean operation to read out graph representations, followed by an MLP layer. 

\textbf{Scalability analysis.} \textcolor{black}{As AutoGCN realizes a simple form in spatial domain without eigendecomposition, the computation complexity of AutoGCN is with the same magnitude as GCN \cite{kipf2017semi}, i.e. $O(M)$, where $M$ is the number of edges. Concretely,  the computation complexity of the low-pass, middle-pass, high-pass filters of AutoGCN is $O(M)$,$O(M)$,$O(M)$ respectively. Therefore, the overall computation complexity of AutoGCN is $O(M)$.}

Overall, the benefit of AutoGCN can be understood in three aspects.  First, AutoGCN captures the full spectrum of graph signals with a minimal set of graph convolutional filters, namely a low-pass filter, a high-pass filter, and a middle-pass filter.  This enhances model expressivity compared to GCNs which only contain a low-pass filter. Second, AutoGCN is able to adjust the bandwidth and magnitude of its filter functions adaptively based on data. It could be extremely helpful when the distribution of data lies far from our prior knowledge. Third, AutoGCN achieves a simple form in the spatial domain.  The low-pass filter can be considered as smoothing a node's self-information with its neighborhood information,  the high-pass filter can be regarded as obtaining the difference between a node's self-information and its neighborhood information, and the middle-pass filter can be taken as distinguishing a node's self-information from its 2nd-hop neighborhood information. All three filters enrich the feature representations of nodes in a different way. 

\begin{table*}[ht]
	\centering
	\caption{Summary of node classification datasets. The number of training graphs, validiation graphs, and test graphs for PUBMED, Arxiv-year, YelphChi, and Squirrel are missing because they only contain a single graph.}
	\label{tab:ncftab}
		\resizebox{1\textwidth}{!}{
	\begin{tabular}{llllllll}
		\toprule
		& \# Nodes & \# Edges & \# Features & \# Class & \# Train Graphs & \# Val Graphs & \# Test Graphs  \\
		\hline
		PUBMED & 19717 & 44338 & 500 & 7 & - & - & -  \\
		SBM-PATTERN    & 50-180 &  4749 on average    &       3      & 2   & 10000           & 2000          & 2000              \\
		SBM-CLUSTER   & 40-190 & 4302 on average    &      7       & 6    & 10000           & 1000          & 1000                \\
		Arxiv-year &  169343 &  1166243  & 128 & 5 & - & - & - \\
		YelpChi & 45954 & 3846979 & 32 & 2 & - & - & - \\
		Squirrel & 5201 & 216933 & 2089 & 5 & -& - & - \\
		\bottomrule
	\end{tabular}}
\end{table*}
 
\section{Experiments}
\label{sec:exp}

We implement AutoGCN using PyTorch within a graph neural network benchmarking framework \cite{dwivedi2020benchmarkgnns} based on DGL \cite{wang2019dgl}.  As pointed out by Dwivedi et al. \cite{dwivedi2020benchmarkgnns}, previous popular graph datasets such as Cora and Tu datasets are small and more likely to be overfitted, making it hard to identify the contribution of new methods.  In this paper, we choose medium-scale graph datasets. To validate the effectiveness of our AutoGCN, we test its performance on node and graph prediction tasks.

\subsection{Node Classification}
\textbf{Datasets.} \textcolor{black}{We use six datasets: PUBMED, SBM-PATTERN, SBM-CLUSTER, Arxiv-year, YelpChi and Squirrel. PUBMED is a citation network consisting of 19,717 documents with 7 class labels.} The feature of each document is a bag-of-words representation of dimension 500.  We adopt the preprocessed version of PUBMED from DGL \cite{wang2019dgl}. SBM-PATTERN and SBM-CLUSTER are synthesized graph datasets \cite{dwivedi2020benchmarkgnns}.  They are generated by the stochastic block model with a probability $p$ that two nodes are connected if they belong to the same community and a probability $q$ that two nodes are linked if they fall into different communities. The node features of  SBM-PATTERN and SBM-CLUSTER  are uniformly generated from a vocabulary of $\{1,2,3\}$. The task of SBM-PATTERN is to predict whether a node belongs to a graph pattern. The task of SBM-CLUSTER is to classify nodes to their belonged clusters.  \textcolor{black}{For Arxiv-year, YelpChi and Squirrel, they are three non-homophilous graph datasets adopted from \cite{lim2021new}. In a non-homophilous graph, connected nodes are not evidently more likely to share the same label. The summary statistics for these three datasets are provided in Table \ref{tab:ncftab}.}

\textbf{Experimental Settings.}
The training loss for all three datasets is the cross-entropy loss.  We report the averaged accuracy on test data over 5 runs with 5 different seeds. \textcolor{black}{For PUBMED, in each run, the dataset is randomly split into train, valid, and test data by 60\%, 20\%, 20\%.  For SBM-PATTERN and SBM-CLUSTER, we follow the same data split as \cite{dwivedi2020benchmarkgnns}. For Arxiv-year, YelpChi, and Squirrel, we follow \cite{lim2021new} to split the datasets into train, valid, and test data with a ratio of 2:1:1. We utilize the validation set to select the best model among all epochs in each run.  For PUBMED, the default hyperparameter setting is used for each baseline model. For SBM-PATTERN, we choose 4 layers for all methods with a fixed budget of around 100k parameters.  For SBM-CLUSTER, we set 8 layers for all methods with a fixed budget of around 200k parameters. For Arxiv-year, YelpChi, and Squirrel, we set 2 layers for all methods with a fixed budget of around 100k parameters. Residual connections, batch normalizations, and graph size normalizations \cite{dwivedi2020benchmarkgnns} are employed for all methods on SBM-PATTERN SBM-CLUSTER, Arxiv-year, Yelp-Chi, and Squirrel. \textcolor{black}{This ensures all baseline methods do not face the over-smoothing problem when the model is deep.} Other hyperparameter settings are provided in Appendix B. }
All experiments are conducted using a single Titan XP GPU card.

\textbf{Baselines.}  We compare AutoGCN against nine methods: MLP, ChebNet \cite{defferrard2016convolutional}, GCN \cite{kipf2017semi}, SGC \cite{wu2019simplifying}, DSGCN \cite{balcilar2020bridging}, APGCN \cite{spinelli2020adaptive}, GIN \cite{xu2018how}, GraphSage \cite{hamilton2017inductive}, GAT \cite{velickovic2018graph}, and MoNet \cite{monti2017geometric}.  We use the DGL built-in implementations of graph convolutional layers for GCN, GraphSage, GAT and MoNet. For MLP and GIN, we adopt the implementations provided by Dwivedi et al. \cite{dwivedi2020benchmarkgnns}.
For ChebNet, we implement it with a second-order approximation and set $\lambda_{max}=2$. 
For DSGCN, we use its original TensorFlow implementation.  Due to DSGCN does not provide a general rule to design customized graph convolutional filters for new datasets, we only test DSGCN on the PUBMED dataset with its default setting.

\textbf{Results \& Discussion.} Table \ref{tab:nresults} shows the experimental results for node classification tasks.  \textcolor{black}{On all datasets except SBM-PATTERN, AutoGCN significantly outperforms baseline methods that only work as a low-pass filter including GCN, SGC, GraphSage, and GAT. We observe that the experimental result of APGCN on SBM-PATTERN outperforms other baseline method. However, the performance of APGCN is not robust across all datasets.} Besides, we observe that DSGCN does not perform well as expected on the PUBMED dataset. The same phenomenon also applies to GAT. In its original paper, GAT evidently outperforms GCN on PUBMED. However, our experiment shows a different conclusion.  This is partly due to the original data split of PUBMED used by GAT and DSGCN is not big enough to evaluate graph neural network models. The original data split consists of 140 nodes for training, 500 nodes for validation, and 1000 nodes for testing \cite{kipf2017semi}.  Therefore, it may easily drive the design of graph neural networks to overfit the small dataset. 

\begin{table*}
\centering
\caption{Performance on node classification tasks. Results on test sets are averaged over 5 runs with 5 different seeds. OOM stands for out of memory. }
\label{tab:nresults}
\resizebox{1\textwidth}{!}{
\begin{tabular}{ l | l  l  l | l  l  l | l  l  l}
\toprule
\multirow{3}{*}{\textbf{Model}} &  \multicolumn{3}{c}{\textbf{PUBMED}} & \multicolumn{3}{c}{\textbf{SBM-PATTERN}} & \multicolumn{3}{c}{\textbf{SBM-CLUSTER}} \\
\cline{2-10}
\\[-1em] 
\\[-1em]
\\[-1em]  
& \textbf{ACC} & \textbf{\#Param}  & \textbf{s/epoch} & \textbf{ACC} & \textbf{\#Param}  & \textbf{s/epoch} & \textbf{ACC} & \textbf{\#Param}  &  \textbf{s/epoch}\\
\midrule
\midrule
\\[-1em] 
	MLP & 0.493 $\pm$ 0.164 & 9019 & 0.0074 &  0.505 $\pm$ 0.000 & 108112 &  7.6688  & 0.223 $\pm$ 0.003 &  201900 & 10.0579 \\ \hline
\\[-1em] 
\\[-1em] 
GIN \cite{xu2018how} & 0.871 $\pm$ 0.004 & 16161 & 0.0097 & 0.863 $\pm$ 0.000 & 101764 & 17.7936 & 0.663 $\pm$ 0.010  & 207412 & 16.7895 \\ \hline
\\[-1em] 
\\[-1em] 
GraphSage  \cite{hamilton2017inductive} &0.868 $\pm$ 0.006 & 16134 & 0.0126 & 0.663 $\pm$  0.002 & 106563 & 12.6709  & 0.666 $\pm$ 0.018 & 205675 & 15.1923 \\ \hline
\\[-1em] 
\\[-1em] 
GAT    \cite{velickovic2018graph} &0.856 $\pm$ 0.005 & 32460 & 0.0285 &  0.780 $\pm$ 0.004 & 109936 &  29.5291 & 0.675 $\pm$ 0.006  &205548 & 39.0724 \\ \hline
\\[-1em] 
\\[-1em] 
MoNet \cite{monti2017geometric} & 0.847 $\pm$ 0.006 & 9903 & 3.6768 & 0.864 $\pm$ 0.000 & 104135 & 856.5839 & 0.663 $\pm$ 0.009 & 203299 & 803.4355 \\
\hline
\\[-1em] 
\\[-1em]  
ChebNet \cite{defferrard2016convolutional} &0.886 $\pm$ 0.006 & 24201 & 0.0226 &  0.857 $\pm$ 0.000 & 103703 & 15.7381  & 0.734 $\pm$ 0.004   & 202435 & 15.4712 \\ \hline
\\[-1em] 
\\[-1em] 
GCN  \cite{kipf2017semi}  & 0.866 $\pm$ 0.005  & 8105 & 0.0112  & 0.855 $\pm$ 0.000  & 106463 & 12.4331  & 0.645 $\pm$ 0.007 & 199015  & 14.7601 \\ \hline
\\[-1em] 
\\[-1em] 
\textcolor{black}{SGC} \cite{wu2019simplifying} & 0.866 $\pm$ 0.004 & 8067 & 0.0049 & 0.835 $\pm$ 0.001  & 108112 & 7.2495 & 0.475 $\pm$ 0.001 & 201900 & 5.7526  \\ \hline
\\[-1em] 
\\[-1em] 
\textcolor{black}{APGCN} \cite{spinelli2020adaptive} & 0.851 $\pm$ 0.005 & 32263 & 0.0447 & \textbf{1.000 $\pm$ 0.000} & 106124 & 7.7181 &  0.345 $\pm$ 0.002 & 202868 & 13.0703 \\ \hline
\\[-1em] 
\\[-1em] 

DSGCN \cite{balcilar2020bridging} & 0.853 $\pm$ 0.007& 9080 & 19.7565 &-&-&-&-&-&-\\ \hline
\\[-1em] 
\\[-1em]
\textbf{AutoGCN} & \textbf{0.893 $\pm$ 0.005} & 24297 & 0.0372 & 0.859 $\pm$ 0.000 & 103895 & 20.7730 &  \textbf{0.741 $\pm$ 0.002} & 202819 & 20.3951 \\  

\midrule
\midrule

\multirow{3}{*}{\textbf{Model}} &  \multicolumn{3}{c}{\textbf{Arxiv-year}} & \multicolumn{3}{c}{\textbf{YelpChi}} & \multicolumn{3}{c}{\textbf{Squirrel}} \\
\cline{2-10}
\\[-1em] 
\\[-1em]
\\[-1em]  
& \textbf{ACC} & \textbf{\#Param}  & \textbf{s/epoch} & \textbf{ACC} & \textbf{\#Param}  & \textbf{s/epoch} & \textbf{ACC} & \textbf{\#Param}  &  \textbf{s/epoch}\\
\midrule
\midrule
\\[-1em] 
	MLP & 0.346 $\pm$ 0.001 & 107695 & 0.0525 &  0.854 $\pm$ 0.006 & 102062 & 0.0128 & 0.195 $\pm$ 0.006 & 104213 & 0.0041 \\ \hline
\\[-1em] 
\\[-1em] 
GIN \cite{xu2018how} & 0.440 $\pm$ 0.003 & 104578 & 0.1200 & 0.855 $\pm$ 0.005 & 105013 & 0.0553 & 0.265 $\pm$ 0.024 & 101908 & 0.0111 \\ \hline
\\[-1em] 
\\[-1em] 
GraphSage  \cite{hamilton2017inductive} &  0.446 $\pm$ 0.002 & 103710 & 0.0984 & 0.854 $\pm$ 0.006 & 103324 & 0.0757 & 0.330 $\pm$ 0.018 & 101770 & 0.0118 \\ \hline
\\[-1em] 
\\[-1em] 
GAT    \cite{velickovic2018graph} & 0.297 $\pm$ 0.017 & 101652 & 0.2460 & OOM & OOM & OOM & 0.272 $\pm$ 0.008 & 103220 & 0.0159 \\ \hline
\\[-1em] 
\\[-1em] 
MoNet \cite{monti2017geometric} & 0.408 $\pm$ 0.016 & 103522 & 28.8757 & OOM & OOM & OOM  & 0.277 $\pm$ 0.053 & 107921 & 5.2633  \\
\hline 
\\[-1em] 
\\[-1em]  
ChebNet \cite{defferrard2016convolutional} & 0.483 $\pm$ 0.001 & 101007 & 0.1198 & 0.880 $\pm$ 0.004 & 105066 & 0.1189 & 0.297 $\pm$ 0.006 & 101391 & 0.0251  \\ \hline
\\[-1em] 
\\[-1em] 
GCN  \cite{kipf2017semi}  & 0.452 $\pm$ 0.001 & 101135 & 0.0784 & 0.854 $\pm$ 0.005 & 102006 & 0.0569 & 0.273 $\pm$ 0.013 & 103119 & 0.0069 \\ \hline
\\[-1em] 
\\[-1em] 
\textcolor{black}{SGC} \cite{wu2019simplifying} &  0.399 $\pm$ 0.001 & 109355 & 0.1060 &  0.854 $\pm$ 0.005 & 100802 & 0.0826 & 0.253 $\pm$ 0.009 & 102917 & 0.0047 \\ \hline
\\[-1em] 
\\[-1em] 
\textcolor{black}{APGCN} \cite{spinelli2020adaptive} & 0.325 $\pm$ 0.006 & 100107 & 0.1137 & 0.854 $\pm$ 0.006 & 100805 & 0.1515 &  0.195 $\pm$ 0.013 & 102923 & 0.0260 \\ \hline
\\[-1em] 
\\[-1em] 
DSGCN \cite{balcilar2020bridging} & - & - & - & - & - & - & - & - & - \\ \hline
\\[-1em] 
\\[-1em]
\textbf{AutoGCN} & \textbf{0.485 $\pm$ 0.001} & 101151 & 0.1761 & \textbf{0.888 $\pm$ 0.006} & 105210 & 0.1504 & \textbf{0.337 $\pm$ 0.026} & 101607 & 0.0364 \\  

\toprule
\end{tabular}
}
\end{table*}




\subsection{Graph Prediction}
\textbf{Datasets.}  We use three medium-scale datasets for graph prediction \cite{dwivedi2020benchmarkgnns}: ZINC, MNIST, and CIFAR10.  ZINC is a molecular graph dataset. The task is to regress a molecular property called the constrained solubility \cite{jin2018junction}.  For each molecular graph, node features are the type of atoms.  MNIST and CIFAR10 are image classification datasets from computer vision. They are converted from images to graphs by representing nodes using super-pixels and constructing edges using k-nearest neighbors.  The summary of statistics for these three datasets is provided in Table \ref{tab:gcftab}.

\begin{table*}
	\centering
	\caption{Summary of graph prediction datasets.}
	\label{tab:gcftab}
	\resizebox{.9\textwidth}{!}{
	\begin{tabular}{lllllll}
		\toprule
		& \# Train Graphs & \# Val Graphs & \# Test Graphs & \# Nodes & \# Features & \# Class \\
		\hline
		Zinc    & 10000           & 1000          & 100            & 9-37     &       28      & -       \\
		MNIST   & 55000           & 5000          & 10000          & 40-75    &     3        & 10       \\
		CIFAR10 & 45000           & 5000          & 10000          & 85-150   &    5         & 10      \\
		\bottomrule
	\end{tabular}}
\end{table*}

\textbf{Experimental Settings.}
The training loss and the evaluation metric for ZINC data is the mean absolute error (MAE). For MNIST and CIFAR10, the training objective is the cross-entropy loss, and the evaluation metric is the accuracy score.  We report the averaged evaluation metrics over test data over 5 runs with 5 different seeds. \textcolor{black}{For all three datasets, we follow the same data split as \cite{dwivedi2020benchmarkgnns}. We utilize the validation set to select the best model among all epochs in each run.  
The same output layer is employed for all baseline methods. It reads out the graph representation by taking the average over all node features from the last layer and passes it to an MLP.  For all three datasets, we set 8 layers for all methods with a fixed budget of around 200k parameters. Residual connections, batch normalizations, and graph size normalizations are employed for all methods on all three datasets. Other hyperparameter settings can be found in Appendix B. } All experiments are conducted using a single Titan XP GPU card.

\textbf{Results \& Discussion.} Table \ref{tab:graphc} shows the experimental results of graph classification tasks. According to Table \ref{tab:graphc},  AutoGCN outperforms baseline methods on all three datasets.
The effectiveness of the introduction of automatic graph convolutional filter functions can be inferred from the comparison between ChebNet and AutoGCN. In fact, the graph convolutional kernels of both methods contain a zeroth-order, first-order, and second-order of the graph Laplacian matrix. Correspondingly, ChebNet consists of an all-pass filter, a low-high-pass filter, and a middle-pass filter. The main difference lies in that AutoGCN adjusts frequency profiles of graph convolutional filters adaptively according to data inputs.  Though the performance difference between AutoGCN and ChebNet is smaller than the gap between AutoGCN and other baseline methods,  according to Table \ref{tab:nresults} and Table \ref{tab:graphc},  AutoGCN consistently outperforms ChebNet over all datasets, showing the effectiveness of our proposed method. 

\textbf{Baselines.} The choice of baseline methods is the same as that in the node classification task.
\begin{table*}
	\centering
	\caption{Performance on graph prediction tasks. Results on test sets are averaged over 5 runs with 5 different seeds.}
	\label{tab:graphc}
	\resizebox{1\textwidth}{!}{
		\begin{tabular}{ l | l  l  l | l  l  l | l  l  l}
			\toprule
			\multirow{3}{*}{\textbf{Model}} &  \multicolumn{3}{c}{\textbf{ZINC}} & \multicolumn{3}{c}{\textbf{MINST}} & \multicolumn{3}{c}{\textbf{CIFAR10}} \\
			\cline{2-10}
			\\[-1em]  
			\\[-1em]  
			\\[-1em] 
			& \textbf{MAE} & \textbf{\#Param}  & \textbf{s/epoch} & \textbf{ACC} & \textbf{\#Param}  & \textbf{s/epoch} & \textbf{ACC} & \textbf{\#Param}  &  \textbf{s/epoch}\\
			\midrule
			\midrule
			\\[-1em] 
			\\[-1em] 
			MLP & 0.713 $\pm$ 0.001  & 204897 & 1.6209  & 0.951 $\pm$ 0.003 & 202383 & 36.6067 & 0.536 $\pm$ 0.007 & 202707 & 48.5513 \\ \hline
			\\[-1em] 
			\\[-1em] 
			GIN \cite{xu2018how} & 0.249 $\pm$ 0.008  &204727 & 5.3311 & 0.974  $\pm$  0.001 & 211078 & 59.0530 & 0.651 $\pm$ 0.005 & 211298 & 74.2930 \\ \hline
			\\[-1em] 
			\\[-1em] 
			GraphSage  \cite{hamilton2017inductive} & 0.378 $\pm$ 0.015  & 207845 & 3.5253  & 0.977 $\pm$ 0.001& 205457 & 45.2319  & 0.669 $\pm$ 0.002  & 205677 & 57.7408 \\ \hline
			\\[-1em] 
			\\[-1em] 
			GAT    \cite{velickovic2018graph}  &  0.424 $\pm$ 0.006  &   208545   & 5.6672 & 0.966 $\pm$ 0.001 & 205248 & 70.3261 & 0.632 $\pm$ 0.008 & 205552 & 92.0455 \\ \hline
			\\[-1em] 
			\\[-1em] 
			MoNet \cite{monti2017geometric} & 0.340 $\pm$ 0.008 &205074 & 8.9661 & 0.942 $\pm$ 0.004 & 203121 & 655.3304 & 0.623 $\pm$ 0.003 & 203301 & 926.6127 \\
			\hline
			\\[-1em] 
			\\[-1em] 
			ChebNet  \cite{defferrard2016convolutional}    & 0.254 $\pm$ 0.008 & 204210   &  4.4090  &  0.974 $\pm$ 0.002   &  202257 & 42.5235  & 0.677 $\pm$ 0.004 & 202437 & 63.5918  \\ \hline
			\\[-1em] 
			\\[-1em] 
			GCN  \cite{kipf2017semi}    & 0.314 $\pm$ 0.009 &  201975    &  3.2912   & 0.911 $\pm$ 0.002 & 198717 & 47.9303 & 0.551 $\pm$ 0.004 & 199017 & 57.0702\\ \hline
			\\[-1em] 
			\\[-1em] 

			SGC \cite{wu2019simplifying} & 0.725 $\pm$ 0.005 & 199575 & 1.0961 & 0.956 $\pm$ 0.002 & 196317 & 21.4136 & 0.542 $\pm$ 0.003 & 196617 & 21.8775  \\
		    \\[-1em] 
			\\[-1em] 
			\hline
			APGCN \cite{spinelli2020adaptive} & 0.819 $\pm$ 0.004 & 199726 & 2.2744 &  0.958 $\pm$ 0.002 & 204766 & 39.6259 &  0.523 $\pm$ 0.009 & 205056  & 67.6515\\
			\hline

			\\[-1em] 
			\\[-1em] 

			\textbf{AutoGCN}  & \textbf{0.225 $\pm$ 0.006}     &  204594 & 4.8771 & \textbf{0.978 $\pm$ 0.001} &  202641 & 59.3858  & \textbf{0.678 $\pm$ 0.002} & 202821 & 77.5203  \\  
			\toprule
		\end{tabular}
	}
\end{table*}

\subsection{Ablation Study}
We conduct an ablation study on SBM-CLUSTER and ZINC to validate the effectiveness of key components that contribute to the improvement of our proposed model. We name AutoGCN without different components as follows - \textbf{w/o low}: AutoGCN without the low-pass linear filter; \textbf{w/o high}: AutoGCN without the high-pass linear filter; \textbf{w/o middle}: AutoGCN without the middle-pass quadratic filter; \textbf{w/o over}: AutoGCN  without over-parameterizing  filter functions; \textbf{w/o par}: AutoGCN without parameterizing  filter functions. We set $p=1,a=0.5$ as fixed values for the low-pass filter, the high-pass filter, and the middle-pass filter; \textbf{w/o gate}: AutoGCN without the complementary gate. We simply sum up the outputs of three graph convolutional filters in each layer. For \textbf{w/o low}, \textbf{w/o high}, and \textbf{w/o middle}, we increase the hidden feature dimension to keep approximately the same number of parameters as AutoGCN.  We run each experiment five times and report the averaged evaluation metrics with standard deviation on test data. Table \ref{tab:ablation} reports the experimental results of the ablation study.  It shows that any absence of a key component reduces the performance of AutoGCN. 
\subsection{Discussion \& Limitation}
\textcolor{black}{There are close connections between AutoGCN and prior works. Like ChebNet \cite{defferrard2016convolutional} and GCN \cite{kipf2017semi}, AutoGCN is a polynomial approximation on the general filtering function. Similar to DSGCN \cite{balcilar2020bridging} , AutoGCN rely on manually choosing base filtering functions. Moreover, the linear filtering functions of AutoGCN is similar to the residual connection applied many GNN works such as GCNII \cite{chen2020simple}.  The mid-pass filtering function is close to performing 2-hop propagation studied in MixHop \cite{abu2019mixhop}. The fact that AutoGCN introduces different message passing mechanisms for different frequency components can be related to the idea of heterogeneous graph neural networks \cite{wang2019heterogeneous, zhang2019heterogeneous, fu2020magnn} where they tackle different types of relations with different networks. Despite the common characteristics, the distinct difference between AutoGCN and these prior works lies in  two folds. First, it contains a high-frequency filter component. The high-frequency filter is used to distinguish the difference between a node and its neighbors. Second, the frequency bandwidth of AutoGCN is not fixed. It can be adjusted automatically by learning from data.}

\textcolor{black}{While being adaptive to frequency bandwidth, the flexibility of AutoGCN in adjusting filtering functions is limited. The linear and quadratic functions of AutoGCN are only approximating low-pass, mid-pass and high-pass filters. On the one side, each type of filter functions still contain other frequency bands which are not desired. For example, the designed low-pass frequency functions also has a small portion of mid-frequency bands and high-frequency bands.
On the other side, if we choose to cut-off the frequency for the corresponding filters, the model cannot be simplified to first-order and second-order of the adjacency matrix, and thus such alternative model is no longer scalable.
}


\begin{table*}
	\centering
	\caption{Ablation study.}
	\label{tab:ablation}
	\resizebox{\textwidth}{!}{
		\begin{tabular}{l| l|l|l|l|l|l|l}
			\toprule
			Data & AutoGCN & w/o low & w/o high & w/o middle & w/o over   & w/o par & w/o gate  \\
			\\[-1em]  
			\hline
			\\[-1em]
			SBM-CLUSTER & \textbf{74.104 $\pm$ 0.151} & 72.168 $\pm$ 0.232 & 73.748 $\pm$ 0.181 & 71.924 $\pm$ 0.329 & 73.559 $\pm$ 0.283 &  73.450 $\pm$ 0.438 & 73.991 $\pm$ 0.206 \\ 
			\\[-1em]  
			\hline
			\\[-1em]
			ZINC   & \textbf{0.225 $\pm$ 0.009} & 0.253 $\pm$ 0.011 & 0.253 $\pm$ 0.010 & 0.242 $\pm$ 0.015 & 0.233 $\pm$ 0.003 & 0.231 $\pm$ 0.007 & 0.248 $\pm$ 0.006\\
			\bottomrule 
		\end{tabular}
	}

\end{table*}
\begin{figure*}

	\begin{subfigure}[b]{0.35\textwidth}   
		\centering 
		\includegraphics[width=\textwidth]{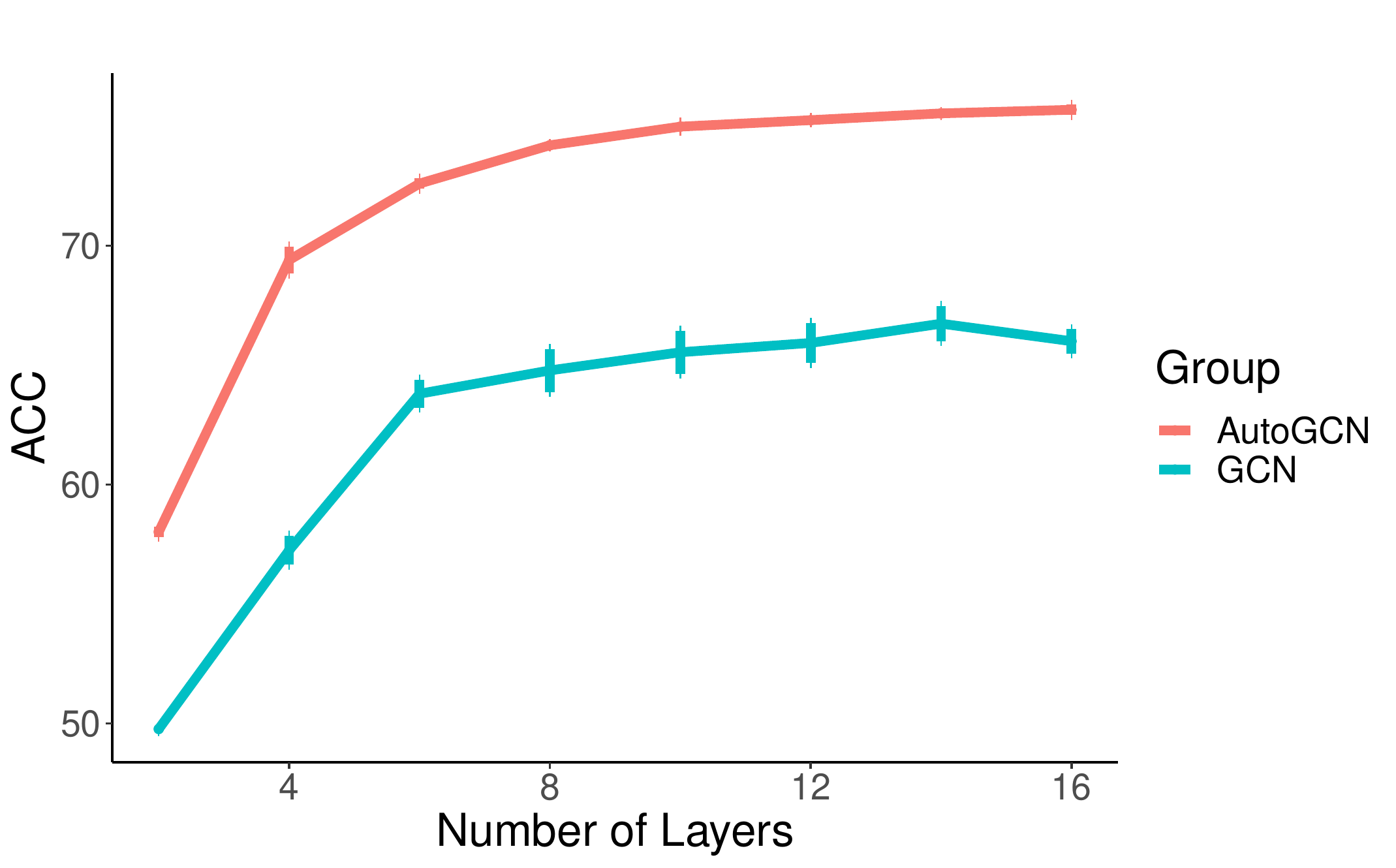}
		\caption[The trend of ACC v.s. model depth for the SBM-CLUSTER dataset. ]
		{{\small The trend of ACC v.s model depth  for the SBM-CLUSTER dataset. }}    
		\label{fig:d1}
	\end{subfigure}
	\hfill
	\begin{subfigure}[b]{0.35\textwidth}   
		\centering 
		\includegraphics[width=\textwidth]{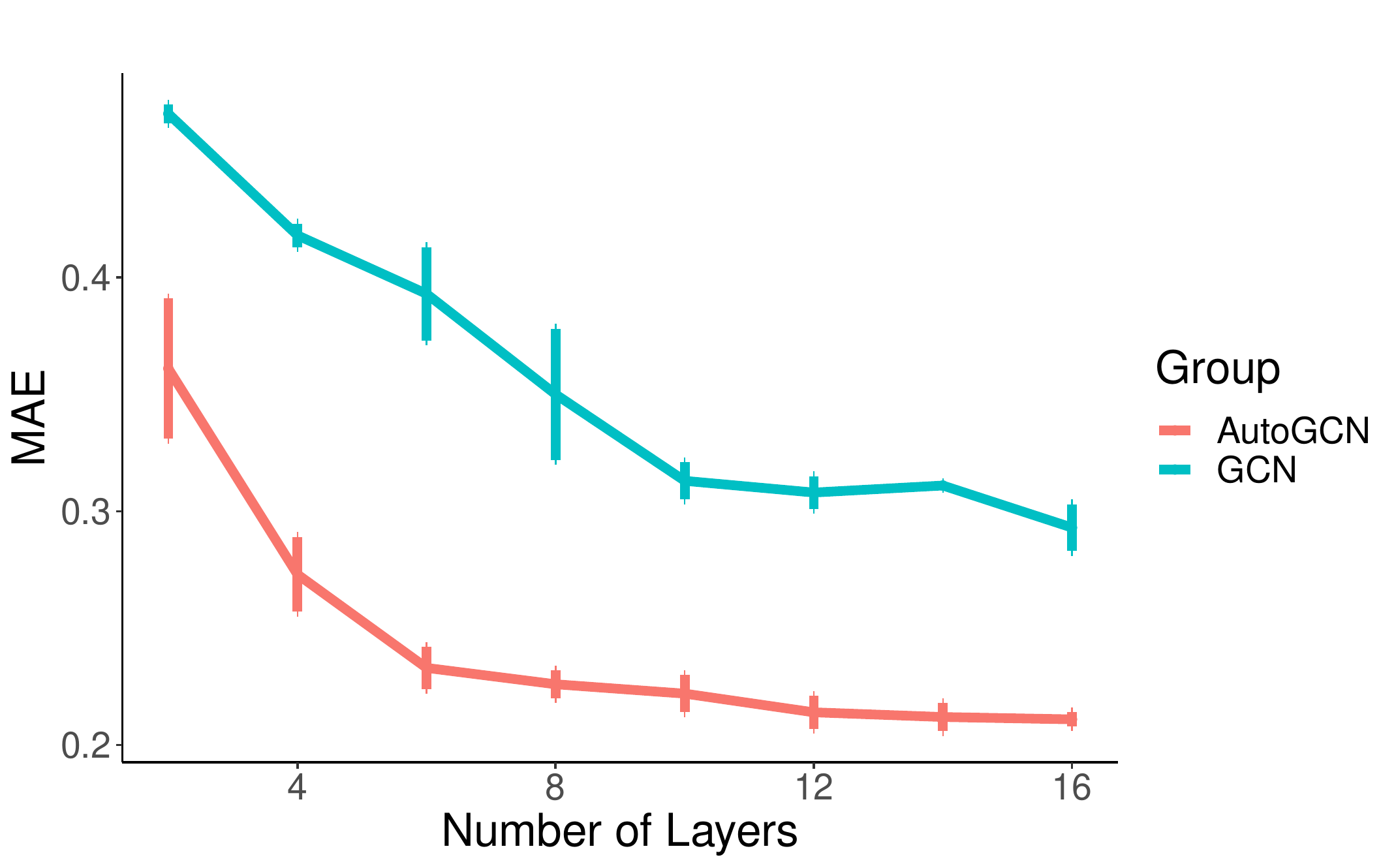}
		\caption[The trend of MAE v.s. model depth for the ZINC dataset. .]	
		{{\small The trend of MAE v.s. model depth for the ZINC dataset. }}    
		\label{fig:d2}
	\end{subfigure}	
	\caption[Comparison analysis of model depth between AutoGCN and GCN.]
	{\small Comparison analysis of model depth between AutoGCN and GCN.} 
	\label{fig:depth}
\end{figure*}

\subsection{Analysis of Model Depth}
Graph convolutional networks often face the over-smoothing problem. In theory, as the number of hidden layers goes to infinity, node features of a graph will converge to a fixed point \cite{li2018deeper}.  A solution to mitigate the over-smoothing problem is to add residual connections to each layer. Figure \ref{fig:depth} plots the performance curve of GCN and AutoGCN on SBM-CLUSTER and ZINC as the number of layers is increased from 2 to 16 every two steps. 
We run each experiment five times and report the averaged evaluation metrics with standard deviation on test data. With the increase in the number of layers, the performance of GCN and AutoGCN is also enhanced. It demonstrates the usefulness of residual connections in designing deep graph convolutional networks. Note that it may be argued that the significant improvement of AutoGCN over GCN is mainly attributed to the fact that each layer of AutoGCN receives a broader range of neighborhood information because of the second-order of the graph Laplacian matrix. It is suspected that comparing AutoGCN with GCN or other methods under the same number of layers may not be a fair option. However, if we look at Figure \ref{fig:depth},  a deeper GCN  still cannot compete with a shallower AutoGCN.  For example,  a four-layer GCN performs much worse than a two-layer AutoGCN
, even the depth of node information aggregation for both models is 4. This may suggest that the second-order graph Laplacian matrix not only broadens a node's neighborhood but also works as a middle-pass filter to filter low and high frequency of graph signals.


\section{Conclusion}
\label{sec:conclusion}

This paper proposes a graph convolutional network that captures the full spectrum of graph signals with automatic filtering. Our method AutoGCN consists of three different forms of graph convolutional filters: the low-pass filter, the high-pass filter, and the middle-pass filter. All three filters enrich node feature representations in a different way. In each filter, we introduce two parameters to control the magnitude and the curvature of its frequency profile. It enables AutoGCN to update its parameterized filter functions based on data. Our experiments show that AutoGCN achieves significant improvement over baseline methods that only work as low-pass filters. Different graph convolutional filters have great potential to deal with graphs with heterophily \cite{zheng2022graph}, i.e., nodes with different labels tend to be linked. In the future, we plan to apply our method to heterophilic graphs and exploit the trustworthiness \cite{zhang2022trustworthy} of our model (e.g., robustness).


\bibliographystyle{IEEEtran}
\bibliography{cite.bib}

\begin{thebibliography}{10}
\providecommand{\url}[1]{#1}
\csname url@samestyle\endcsname
\providecommand{\newblock}{\relax}
\providecommand{\bibinfo}[2]{#2}
\providecommand{\BIBentrySTDinterwordspacing}{\spaceskip=0pt\relax}
\providecommand{\BIBentryALTinterwordstretchfactor}{4}
\providecommand{\BIBentryALTinterwordspacing}{\spaceskip=\fontdimen2\font plus
\BIBentryALTinterwordstretchfactor\fontdimen3\font minus
  \fontdimen4\font\relax}
\providecommand{\BIBforeignlanguage}[2]{{%
\expandafter\ifx\csname l@#1\endcsname\relax
\typeout{** WARNING: IEEEtran.bst: No hyphenation pattern has been}%
\typeout{** loaded for the language `#1'. Using the pattern for}%
\typeout{** the default language instead.}%
\else
\language=\csname l@#1\endcsname
\fi
#2}}
\providecommand{\BIBdecl}{\relax}
\BIBdecl

\bibitem{bronstein2017geometric}
M.~M. Bronstein, J.~Bruna, Y.~LeCun, A.~Szlam, and P.~Vandergheynst,
  ``Geometric deep learning: going beyond euclidean data,'' \emph{IEEE Signal
  Processing Magazine}, vol.~34, no.~4, pp. 18--42, 2017.

\bibitem{wu2020comprehensive}
Z.~Wu, S.~Pan, F.~Chen, G.~Long, C.~Zhang, and S.~Y. Philip, ``A comprehensive
  survey on graph neural networks,'' \emph{IEEE Transactions on Neural Networks
  and Learning Systems}, 2020.

\bibitem{zhang2020deep}
Z.~Zhang, P.~Cui, and W.~Zhu, ``Deep learning on graphs: A survey,'' \emph{IEEE
  Transactions on Knowledge and Data Engineering}, 2020.

\bibitem{liu2021graph}
Y.~Liu, M.~Jin, S.~Pan, C.~Zhou, Y.~Zheng, F.~Xia, and P.~S. Yu, ``Graph
  self-supervised learning: A survey,'' \emph{IEEE Transactions on Knowledge
  and Data Engineering}, 2022.

\bibitem{gilmer2017neural}
J.~Gilmer, S.~S. Schoenholz, P.~F. Riley, O.~Vinyals, and G.~E. Dahl, ``Neural
  message passing for quantum chemistry,'' in \emph{Proceedings of the 34th
  International Conference on Machine Learning}, 2017, pp. 1263--1272.

\bibitem{wang2017mgae}
C.~Wang, S.~Pan, G.~Long, X.~Zhu, and J.~Jiang, ``Mgae: Marginalized graph
  autoencoder for graph clustering,'' in \emph{Proceedings of the 26th ACM
  International Conference on Information and Knowledge Management}.\hskip 1em
  plus 0.5em minus 0.4em\relax ACM, 2017, pp. 889--898.

\bibitem{zhang2018link}
M.~Zhang and Y.~Chen, ``Link prediction based on graph neural networks,'' in
  \emph{Proceedings of the 32nd Conference on Neural Information Processing
  Systems}, 2018.

\bibitem{simonovsky2018graphvae}
M.~Simonovsky and N.~Komodakis, ``Graphvae: Towards generation of small graphs
  using variational autoencoders,'' in \emph{Proceedings of the 27th
  International Conference on Artificial Neural Networks}.\hskip 1em plus 0.5em
  minus 0.4em\relax Springer, 2018, pp. 412--422.

\bibitem{pan2019learning}
S.~Pan, R.~Hu, S.-f. Fung, G.~Long, J.~Jiang, and C.~Zhang, ``Learning graph
  embedding with adversarial training methods,'' \emph{IEEE Transactions on
  Cybernetics}, vol.~50, no.~6, pp. 2475--2487, 2019.

\bibitem{wang2020graph}
H.~Wang, C.~Zhou, X.~Chen, J.~Wu, S.~Pan, and J.~Wang, ``Graph stochastic
  neural networks for semi-supervised learning,'' \emph{Advances in Neural
  Information Processing Systems}, vol.~33, 2020.

\bibitem{wu2020openwgl}
M.~Wu, S.~Pan, and X.~Zhu, ``Openwgl: Open-world graph learning,'' in
  \emph{Proc. Of the 20th IEEE International Conference on Data Mining,
  November 17-20, 2020, Sorrento, Italy}, 2020.

\bibitem{zhu2020graph}
S.~Zhu, S.~Pan, C.~Zhou, J.~Wu, Y.~Cao, and B.~Wang, ``Graph geometry
  interaction learning,'' \emph{Advances in Neural Information Processing
  Systems}, vol.~33, 2020.

\bibitem{ying2018graph}
R.~Ying, R.~He, K.~Chen, P.~Eksombatchai, W.~L. Hamilton, and J.~Leskovec,
  ``Graph convolutional neural networks for web-scale recommender systems,'' in
  \emph{Proceedings of the 24th ACM SIGKDD International Conference on
  Knowledge Discovery and Data Mining}.\hskip 1em plus 0.5em minus 0.4em\relax
  ACM, 2018, pp. 974--983.

\bibitem{marcheggiani2017encoding}
D.~Marcheggiani and I.~Titov, ``Encoding sentences with graph convolutional
  networks for semantic role labeling,'' in \emph{Proceedings of the 2017
  Conference on Empirical Methods in Natural Language Processing}, 2017, pp.
  1506--1515.

\bibitem{wang2018dynamic}
Y.~Wang, Y.~Sun, Z.~Liu, S.~E. Sarma, M.~M. Bronstein, and J.~M. Solomon,
  ``Dynamic graph cnn for learning on point clouds,'' \emph{ACM Transactions on
  Graphics (TOG)}, 2019.

\bibitem{wu2019graph}
Z.~Wu, S.~Pan, G.~Long, J.~Jiang, and C.~Zhang, ``Graph wavenet for deep
  spatial-temporal graph modeling,'' in \emph{Proceedings of the 28th
  International Joint Conference on Artificial Intelligence}, 2019.

\bibitem{wu2020connecting}
Z.~Wu, S.~Pan, G.~Long, J.~Jiang, X.~Chang, and C.~Zhang, ``Connecting the
  dots: Multivariate time series forecasting with graph neural networks,'' in
  \emph{Proceedings of the 26th ACM SIGKDD International Conference on
  Knowledge Discovery and Data Mining}.\hskip 1em plus 0.5em minus 0.4em\relax
  ACM, 2018, p. 753–763.

\bibitem{balcilar2020bridging}
M.~Balcilar, G.~Renton, P.~H{\'e}roux, B.~Gauzere, S.~Adam, and P.~Honeine,
  ``Bridging the gap between spectral and spatial domains in graph neural
  networks,'' \emph{arXiv preprint arXiv:2003.11702}, 2020.

\bibitem{bruna2013spectral}
J.~Bruna, W.~Zaremba, A.~Szlam, and Y.~LeCun, ``Spectral networks and locally
  connected networks on graphs,'' in \emph{Proceedings of 2nd International
  Conference on Learning Representations}, 2014.

\bibitem{defferrard2016convolutional}
M.~Defferrard, X.~Bresson, and P.~Vandergheynst, ``Convolutional neural
  networks on graphs with fast localized spectral filtering,'' in
  \emph{Proceedings of the 30th Conference on Neural Information Processing
  Systems}, 2016, pp. 3844--3852.

\bibitem{kipf2017semi}
T.~N. Kipf and M.~Welling, ``Semi-supervised classification with graph
  convolutional networks,'' in \emph{Proceedings of the 5th International
  Conference on Learning Representations}, 2017.

\bibitem{li2018adaptive}
R.~Li, S.~Wang, F.~Zhu, and J.~Huang, ``Adaptive graph convolutional neural
  networks,'' in \emph{Proceedings of the 32nd AAAI Conference on Artificial
  Intelligence}, 2018, pp. 3546--3553.

\bibitem{zhuang2018dual}
C.~Zhuang and Q.~Ma, ``Dual graph convolutional networks for graph-based
  semi-supervised classification,'' in \emph{Proceedings of the Web
  Conference}, 2018, pp. 499--508.

\bibitem{wu2019simplifying}
F.~Wu, T.~Zhang, A.~H.~d. Souza~Jr, C.~Fifty, T.~Yu, and K.~Q. Weinberger,
  ``Simplifying graph convolutional networks,'' in \emph{Proceedings of the
  36th International Conference on Machine Learning}, 2019.

\bibitem{shuman2013emerging}
D.~I. Shuman, S.~K. Narang, P.~Frossard, A.~Ortega, and P.~Vandergheynst, ``The
  emerging field of signal processing on graphs: Extending high-dimensional
  data analysis to networks and other irregular domains,'' \emph{IEEE Signal
  Processing Magazine}, vol.~30, no.~3, pp. 83--98, 2013.

\bibitem{levie2018cayleynets}
R.~Levie, F.~Monti, X.~Bresson, and M.~M. Bronstein, ``Cayleynets: Graph
  convolutional neural networks with complex rational spectral filters,''
  \emph{IEEE Transactions on Signal Processing}, vol.~67, no.~1, pp. 97--109,
  2018.

\bibitem{hammond2011wavelets}
D.~K. Hammond, P.~Vandergheynst, and R.~Gribonval, ``Wavelets on graphs via
  spectral graph theory,'' \emph{Applied and Computational Harmonic Analysis},
  vol.~30, no.~2, pp. 129--150, 2011.

\bibitem{xu2019graph}
B.~Xu, H.~Shen, Q.~Cao, Y.~Qiu, and X.~Cheng, ``Graph wavelet neural network,''
  in \emph{Proceedings of the 7th International Conference on Learning
  Representations}, 2019.

\bibitem{scarselli2009graph}
F.~Scarselli, M.~Gori, A.~C. Tsoi, M.~Hagenbuchner, and G.~Monfardini, ``The
  graph neural network model,'' \emph{IEEE Transactions on Neural Networks},
  vol.~20, no.~1, pp. 61--80, 2009.

\bibitem{li2015gated}
Y.~Li, D.~Tarlow, M.~Brockschmidt, and R.~Zemel, ``Gated graph sequence neural
  networks,'' in \emph{Proceedings of the 3rd International Conference on
  Learning Representations}, 2015.

\bibitem{dai2018learning}
H.~Dai, Z.~Kozareva, B.~Dai, A.~Smola, and L.~Song, ``Learning steady-states of
  iterative algorithms over graphs,'' in \emph{Proceedings of the International
  Conference on Machine Learning}, 2018, pp. 1114--1122.

\bibitem{micheli2009neural}
A.~Micheli, ``Neural network for graphs: A contextual constructive approach,''
  \emph{IEEE Transactions on Neural Networks}, vol.~20, no.~3, pp. 498--511,
  2009.

\bibitem{atwood2016diffusion}
J.~Atwood and D.~Towsley, ``Diffusion-convolutional neural networks,'' in
  \emph{Proc. of NIPS}, 2016, pp. 1993--2001.

\bibitem{monti2017geometric}
F.~Monti, D.~Boscaini, J.~Masci, E.~Rodola, J.~Svoboda, and M.~M. Bronstein,
  ``Geometric deep learning on graphs and manifolds using mixture model cnns,''
  in \emph{Proceedings of 2017 IEEE Conference on Computer Vision and Pattern
  Recognition}.\hskip 1em plus 0.5em minus 0.4em\relax IEEE, 2017.

\bibitem{velickovic2017graph}
P.~Velickovic, G.~Cucurull, A.~Casanova, A.~Romero, P.~Lio, and Y.~Bengio,
  ``Graph attention networks,'' in \emph{Proceedings of the 5th International
  Conference on Learning Representations}, 2017.

\bibitem{xu2018how}
K.~Xu, W.~Hu, J.~Leskovec, and S.~Jegelka, ``How powerful are graph neural
  networks?'' in \emph{Proceedings of the 7th International Conference on
  Learning Representations}, 2019.

\bibitem{klicpera2018predict}
J.~Klicpera, A.~Bojchevski, and S.~G{\"u}nnemann, ``Predict then propagate:
  Graph neural networks meet personalized pagerank,'' in \emph{Proceedings of
  the 7th International Conference on Learning Representations}, 2019.

\bibitem{you2021identity}
J.~You, J.~Gomes-Selman, R.~Ying, and J.~Leskovec, ``Identity-aware graph
  neural networks,'' \emph{arXiv preprint arXiv:2101.10320}, 2021.

\bibitem{zeng2021decoupling}
H.~Zeng, M.~Zhang, Y.~Xia, A.~Srivastava, A.~Malevich, R.~Kannan, V.~Prasanna,
  L.~Jin, and R.~Chen, ``Decoupling the depth and scope of graph neural
  networks,'' \emph{Advances in Neural Information Processing Systems},
  vol.~34, 2021.

\bibitem{chen2020simple}
M.~Chen, Z.~Wei, Z.~Huang, B.~Ding, and Y.~Li, ``Simple and deep graph
  convolutional networks,'' in \emph{International Conference on Machine
  Learning}.\hskip 1em plus 0.5em minus 0.4em\relax PMLR, 2020, pp. 1725--1735.

\bibitem{gao2019graph}
H.~Gao and S.~Ji, ``Graph u-nets,'' in \emph{Proceedings of the International
  Conference on Machine Learning}, 2019.

\bibitem{luan2019break}
S.~Luan, M.~Zhao, X.-W. Chang, and D.~Precup, ``Break the ceiling: Stronger
  multi-scale deep graph convolutional networks,'' in \emph{Advances in neural
  information processing systems}, 2019, pp. 10\,945--10\,955.

\bibitem{abu2019mixhop}
S.~Abu-El-Haija, B.~Perozzi, A.~Kapoor, N.~Alipourfard, K.~Lerman,
  H.~Harutyunyan, G.~V. Steeg, and A.~Galstyan, ``Mixhop: Higher-order graph
  convolutional architectures via sparsified neighborhood mixing,'' in
  \emph{Proceedings of the International Conference on Machine Learning}, 2019.

\bibitem{coifman2006diffusion}
R.~R. Coifman and M.~Maggioni, ``Diffusion wavelets,'' \emph{Applied and
  Computational Harmonic Analysis}, vol.~21, no.~1, pp. 53--94, 2006.

\bibitem{gao2019geometric}
F.~Gao, G.~Wolf, and M.~Hirn, ``Geometric scattering for graph data analysis,''
  in \emph{International Conference on Machine Learning}.\hskip 1em plus 0.5em
  minus 0.4em\relax PMLR, 2019, pp. 2122--2131.

\bibitem{dwivedi2020benchmarkgnns}
V.~P. Dwivedi, C.~K. Joshi, T.~Laurent, Y.~Bengio, and X.~Bresson,
  ``Benchmarking graph neural networks,'' \emph{arXiv preprint
  arXiv:2003.00982}, 2020.

\bibitem{wang2019dgl}
M.~Wang, L.~Yu, D.~Zheng, Q.~Gan, Y.~Gai, Z.~Ye, M.~Li, J.~Zhou, Q.~Huang,
  C.~Ma, Z.~Huang, Q.~Guo, H.~Zhang, H.~Lin, J.~Zhao, J.~Li, A.~J. Smola, and
  Z.~Zhang, ``Deep graph library: Towards efficient and scalable deep learning
  on graphs,'' in \emph{ICLR Workshop on Representation Learning on Graphs and
  Manifolds}, 2019.

\bibitem{lim2021new}
D.~Lim, X.~Li, F.~Hohne, and S.-N. Lim, ``New benchmarks for learning on
  non-homophilous graphs,'' \emph{arXiv preprint arXiv:2104.01404}, 2021.

\bibitem{spinelli2020adaptive}
I.~Spinelli, S.~Scardapane, and A.~Uncini, ``Adaptive propagation graph
  convolutional network,'' \emph{IEEE Transactions on Neural Networks and
  Learning Systems}, 2020.

\bibitem{hamilton2017inductive}
W.~Hamilton, Z.~Ying, and J.~Leskovec, ``Inductive representation learning on
  large graphs,'' in \emph{Proceedings of the 31st Conference on Neural
  Information Processing Systems}, 2017, pp. 1024--1034.

\bibitem{velickovic2018graph}
P.~Veli{\v{c}}kovi{\'{c}}, G.~Cucurull, A.~Casanova, A.~Romero, P.~Li{\`{o}},
  and Y.~Bengio, ``{Graph Attention Networks},'' in \emph{Proceedings of the
  6th International Conference on Learning Representations}, 2018.

\bibitem{jin2018junction}
W.~Jin, R.~Barzilay, and T.~Jaakkola, ``Junction tree variational autoencoder
  for molecular graph generation,'' in \emph{Proceedings of the 35th
  International Conference on Machine Learning}, 2018.

\bibitem{wang2019heterogeneous}
X.~Wang, H.~Ji, C.~Shi, B.~Wang, Y.~Ye, P.~Cui, and P.~S. Yu, ``Heterogeneous
  graph attention network,'' in \emph{The world wide web conference}, 2019, pp.
  2022--2032.

\bibitem{zhang2019heterogeneous}
C.~Zhang, D.~Song, C.~Huang, A.~Swami, and N.~V. Chawla, ``Heterogeneous graph
  neural network,'' in \emph{Proceedings of the 25th ACM SIGKDD International
  Conference on Knowledge Discovery \& Data Mining}, 2019, pp. 793--803.

\bibitem{fu2020magnn}
X.~Fu, J.~Zhang, Z.~Meng, and I.~King, ``Magnn: Metapath aggregated graph
  neural network for heterogeneous graph embedding,'' in \emph{Proceedings of
  The Web Conference 2020}, 2020, pp. 2331--2341.

\bibitem{li2018deeper}
Q.~Li, Z.~Han, and X.-M. Wu, ``Deeper insights into graph convolutional
  networks for semi-supervised learning,'' in \emph{Proceedings of the 32nd
  AAAI Conference on Artificial Intelligence}, 2018.

\bibitem{zheng2022graph}
X.~Zheng, Y.~Liu, S.~Pan, M.~Zhang, D.~Jin, and P.~S. Yu, ``Graph neural
  networks for graphs with heterophily: A survey,'' \emph{arXiv 2202.07082},
  2022.

\bibitem{zhang2022trustworthy}
H.~Zhang, B.~Wu, X.~Yuan, S.~Pan, H.~Tong, and J.~Pei, ``Trustworthy graph
  neural networks: Aspects, methods and trends,'' \emph{arXiv:2205.07424},
  2022.

\end{thebibliography}

\appendices

\section{Derivation of the graph convolutional  kernels  of low-pass, high-pass, and middle filter functions.}

\theoremstyle{definition}
\begin{definition}[Low-pass linear filter]

The low-pass linear filter function is defined as 
\begin{equation}
F_{low}(\boldsymbol\lambda) = p(1-a\boldsymbol\lambda),
\end{equation}
with $p>0$ and $a\in(0,1)$.

The graph convolutional  kernel of the low-pass filter is then derived as 
\begin{align}
\mathbf{C}_{low} &= \mathbf{U}diag(F_{low}(\boldsymbol\lambda))\mathbf{U}^T\\
& = \mathbf{U}diag(p(1-a\boldsymbol\lambda)\mathbf{U}^T \\
& = p(\mathbf{I}- a\mathbf{L} )\\
& = p(a\mathbf{\tilde{A}}+(1-a)\mathbf{I})
\end{align}

\end{definition}

\begin{table*}
  \begin{center}
    \caption{Hyperparameter settings for all experiments. \textit{L} is the number of layers; \textit{hidden} is the dimension of hidden features; \textit{init lr} is the initial learning rate, \textit{patience} is the decay patience, \textit{min lr} is the stopping lr, \textit{weight decay} is the weight decay rate, all experiments have \textit{lr reduce factor} 0.5.}
    \label{tab:hyperparam}
    \scalebox{0.7}{
    \begin{tabular}{c|l|c|c|c|l|c|c|c|c}
    \toprule
    \multirow{2}{*}{\textbf{Dataset}} & \multirow{2}{*}{\textbf{Model}} & \multicolumn{4}{c|}{\textbf{Hyperparameters}} & \multicolumn{4}{c}{\textbf{Learning Setup}} \\
    \cline{3-10}
    & & L & hidden  & dropout & Other & init lr & patience & min lr & weight decay \\
    \midrule
    \midrule
    
    \parbox[t]{2mm}{\multirow{8}{*}{\rotatebox[origin=c]{90}{PUBMED}}} & MLP & 4 & 16 & 0.5 & - & 1e-2 & - & - & 5e-4 \\
    & GIN & 1 & 16 & 0.5  & n\_mlp\_GIN:2; learn\_eps\_GIN:True; neighbor\_aggr\_GIN:sum & 1e-2 & - & - & 5e-4  \\
    & GraphSage & 1 & 16 & 0.5 & sage\_aggregator:mean & 1e-2 & - & - & 5e-4 \\    
    & GAT & 1  & 8 & 0.6 & n\_heads:8;self\_loop:true & 1e-2 & - & - & 5e-4  \\
    & MoNet & 1 & 16 & 0.5 & kernel:3; pseudo\_dim\_MoNet:2 & 1e-2 & - & - & 5e-4 \\
    & ChebNet & 1 & 16 & 0.5 & K:2 & 1e-2 & - & - & 5e-4\\
    & GCN & 1 & 16 & 0.5 & - & 1e-2 & - & - & 5e-4 \\
    & DSGCN & 1 & 16 & 0.25 & - & 1e-2 & - & - & 5e-4 \\
    & SGC & 1 & 16 & 0.5 & - & 1e-2 & - & - & 5e-4 \\
    & APGCN & 1 & 64 & 0.5 & n\_iter:10; prop\_penalty: 0.05; & 1e-2 & - & -  & 5e-4 \\
    & AutoGCN & 1 & 16 & 0.5 & K:16 & 1e-2 & - & - & 5e-4
    \\
    \midrule
    \parbox[t]{2mm}{\multirow{8}{*}{\rotatebox[origin=c]{90}{SBM-PATTERN}}} & MLP & 4 & 152 & 0 & - & 1e-3 & 5 & 1e-5 & 0 \\
    & GIN & 4 & 110 & 0  & n\_mlp\_GIN:2; learn\_eps\_GIN:True; neighbor\_aggr\_GIN:sum & 1e-3 & 5 & 1e-5 & 0 \\
    & GraphSage & 4 & 110 & 0 & sage\_aggregator:mean & 1e-3 & 5 & 1e-5 & 0 \\    
    & GAT & 4  & 19 & 0 & n\_heads:8;self\_loop:true & 1e-3 & 5 & 1e-5 & 0  \\
    & MoNet & 4 & 90 & 0 & kernel:3; pseudo\_dim\_MoNet:2 & 1e-3 & 5 & 1e-5 & 0 \\
    & ChebNet & 4 & 90 & 0 & K:2 & 1e-3 & 5 & 1e-5 & 0\\
    & GCN & 4 & 150 & 0 & - & 1e-3 & 5 & 1e-5 & 0 \\
    & SGC & 4 & 152 & 0 & - & 1e-2 & 5 & 1e-5 & 0 \\
    & APGCN & 4 & 170 & 0 & n\_iter:10; prop\_penalty: 0.05; & 1e-2 & 5 & 1e-5  & 0 \\
    & AutoGCN & 4 & 90 & 0 & K:16 & 1e-3 & 5 & 1e-5 & 0
    \\
        \midrule
    \parbox[t]{2mm}{\multirow{8}{*}{\rotatebox[origin=c]{90}{SBM-CLUSTER}}} & MLP & 8 & 152 & 0 & - & 1e-3 & 5 & 1e-5 & 0 \\
    & GIN & 8 & 110 & 0  & n\_mlp\_GIN:2; learn\_eps\_GIN:True; neighbor\_aggr\_GIN:sum & 1e-3 & 5 & 1e-5 & 0 \\
    & GraphSage & 8 & 110 & 0 & sage\_aggregator:mean & 1e-3 & 5 & 1e-5 & 0 \\    
    & GAT & 8  & 19 & 0 & n\_heads:8;self\_loop:true & 1e-3 & 5 & 1e-5 & 0  \\
    & MoNet & 8 & 90 & 0 & kernel:3; pseudo\_dim\_MoNet:2 & 1e-3 & 5 & 1e-5 & 0 \\
    & ChebNet & 8 & 90 & 0 & K:2 & 1e-3 & 5 & 1e-5 & 0\\
    & GCN & 8 & 150 & 0 & - & 1e-3 & 5 & 1e-5 & 0 \\
    & SGC & 8 & 152 & 0 & - & 1e-2 & 5 & 1e-5 & 0 \\
    & APGCN & 8 & 162 & 0 & n\_iter:10; prop\_penalty: 0.05; & 1e-2 & 5 & 1e-5  & 0 \\
    & AutoGCN & 8 & 90 & 0 & K:16 & 1e-3 & 5 & 1e-5 & 0
    \\
    \midrule
    \parbox[t]{2mm}{\multirow{8}{*}{\rotatebox[origin=c]{90}{Arxiv-year}}} & MLP & 2 & 220 & 0.5 & - & 1e-2 & - & - & 5e-4 \\
    & GIN & 2 & 170 & 0.5  & n\_mlp\_GIN:2; learn\_eps\_GIN:True; neighbor\_aggr\_GIN:sum & 1e-2 & - & - & 5e-4  \\
    & GraphSage & 2 & 170 & 0.5 & sage\_aggregator:mean & 1e-2 & - & - & 5e-4 \\    
    & GAT & 2  & 32 & 0.5 & n\_heads:8;self\_loop:true & 1e-2 & - & - & 5e-4  \\
    & MoNet & 2 & 115 & 0.5 & kernel:3; pseudo\_dim\_MoNet:2 & 1e-2 & - & - & 5e-4 \\
    & ChebNet & 2 & 128 & 0.5 & K:2 & 1e-2 & - & - & 5e-4\\
    & GCN & 2 & 256 & 0.5 & - & 1e-2 & - & - & 5e-4 \\
    & SGC & 2 & 270 & 0.5 & - & 1e-2 & - & - & 5e-4 \\
    & APGCN & 2 & 256 & 0.5 & n\_iter:10; prop\_penalty: 0.05; & 1e-2 & - & -  & 5e-4 \\
    & AutoGCN & 2 & 128 & 0.5 & K:16 & 1e-2 & - & - & 5e-4
    \\
    \midrule
    \parbox[t]{2mm}{\multirow{8}{*}{\rotatebox[origin=c]{90}{Yelp-chi}}} & MLP & 2 & 240 & 0.5 & - & 1e-2 & - & - & 5e-4 \\
    & GIN & 2 & 210 & 0.5  & n\_mlp\_GIN:2; learn\_eps\_GIN:True; neighbor\_aggr\_GIN:sum & 1e-2 & - & - & 5e-4  \\
    & GraphSage & 2 & 210 & 0.5 & sage\_aggregator:mean & 1e-2 & - & - & 5e-4 \\    
    & GAT & 2  & 32 & 0.5 & n\_heads:8;self\_loop:true & 1e-2 & - & - & 5e-4  \\
    & MoNet & 2 & 115 & 0.5 & kernel:3; pseudo\_dim\_MoNet:2 & 1e-2 & - & - & 5e-4 \\
    & ChebNet & 2 & 170 & 0.5 & K:2 & 1e-2 & - & - & 5e-4\\
    & GCN & 2 & 300 & 0.5 & - & 1e-2 & - & - & 5e-4 \\
    & SGC & 2 & 300 & 0.5 & - & 1e-2 & - & - & 5e-4 \\
    & APGCN & 2 & 300 & 0.5 & n\_iter:10; prop\_penalty: 0.05; & 1e-2 & - & -  & 5e-4 \\    & AutoGCN & 2 & 165 & 0.5 & K:16 & 1e-2 & - & - & 5e-4
    \\
    \midrule
    \parbox[t]{2mm}{\multirow{8}{*}{\rotatebox[origin=c]{90}{Squirrel}}} & MLP & 2 & 48 & 0.5 & - & 1e-2 & - & - & 5e-4 \\
    & GIN & 2 & 24 & 0.5  & n\_mlp\_GIN:2; learn\_eps\_GIN:True; neighbor\_aggr\_GIN:sum & 1e-2 & - & - & 5e-4  \\
    & GraphSage & 2 & 24 & 0.5 & sage\_aggregator:mean & 1e-2 & - & - & 5e-4 \\    
    & GAT & 2  & 6 & 0.5 & n\_heads:8;self\_loop:true & 1e-2 & - & - & 5e-4  \\
    & MoNet & 2 & 45 & 0.5 & kernel:3; pseudo\_dim\_MoNet:2 & 1e-2 & - & - & 5e-4 \\
    & ChebNet & 2 & 16 & 0.5 & K:2 & 1e-2 & - & - & 5e-4\\
    & GCN & 2 & 48 & 0.5 & - & 1e-2 & - & - & 5e-4 \\
    & SGC & 2 & 48 & 0.5 & - & 1e-2 & - & - & 5e-4 \\
    & APGCN & 2 & 48 & 0.5 & n\_iter:10; prop\_penalty: 0.05; & 1e-2 & - & -  & 5e-4 \\    & AutoGCN & 2 & 13 & 0.5 & K:16 & 1e-2 & - & - & 5e-4
    \\
    \midrule
    \parbox[t]{2mm}{\multirow{8}{*}{\rotatebox[origin=c]{90}{ZINC}}} & MLP & 8 & 152 & 0 & - & 1e-3 & 10 & 1e-5 & 0 \\
    & GIN & 8 & 110 & 0  & n\_mlp\_GIN:2; learn\_eps\_GIN:True; neighbor\_aggr\_GIN:sum & 1e-3 & 10 & 1e-5 & 0 \\
    & GraphSage & 8 & 110 & 0 & sage\_aggregator:mean & 1e-3 & 10 & 1e-5 & 0 \\    
    & GAT & 8  & 19 & 0 & n\_heads:8;self\_loop:true & 1e-3 & 10 & 1e-5 & 0  \\
    & MoNet & 8 & 90 & 0 & kernel:3; pseudo\_dim\_MoNet:2 & 1e-3 & 10 & 1e-5 & 0 \\
    & ChebNet & 8 & 90 & 0 & K:2 & 1e-3 & 10 & 1e-5 & 0\\
    & GCN & 8 & 150 & 0 & - & 1e-3 & 10 & 1e-5 & 0 \\
    & SGC & 8 & 150 & 0 & - & 1e-2 & 10 & 1e-5 & 0 \\
    & APGCN & 8 & 150 & 0 & n\_iter:10; prop\_penalty: 0.05; & 1e-2 & 10 & 1e-5  & 0 \\
    & AutoGCN & 8 & 90 & 0 & K:16 & 1e-3 & 10 & 1e-5 & 0
    \\
                \midrule
    \parbox[t]{2mm}{\multirow{8}{*}{\rotatebox[origin=c]{90}{MNIST}}} & MLP & 8 & 162 & 0.1 & - & 1e-3 & 5 & 1e-5 & 0 \\
    & GIN & 8 & 110 & 0.1  & n\_mlp\_GIN:2; learn\_eps\_GIN:True; neighbor\_aggr\_GIN:sum & 1e-3 & 5 & 1e-5 & 0 \\
    & GraphSage & 8 & 110 & 0.1 & sage\_aggregator:mean & 1e-3 & 5 & 1e-5 & 0 \\    
    & GAT & 8  & 19 & 0.1 & n\_heads:8;self\_loop:true & 1e-3 & 5 & 1e-5 & 0  \\
    & MoNet & 8 & 90 & 0.1 & kernel:3; pseudo\_dim\_MoNet:2 & 1e-3 & 5 & 1e-5 & 0 \\
    & ChebNet & 8 & 90 & 0.1 & K:2 & 1e-3 & 5 & 1e-5 & 0\\
    & GCN & 8 & 150 & 0.1 & - & 1e-3 & 5 & 1e-5 & 0 \\
    & SGC & 8 & 150 & 0.1 & - & 1e-2 & 5 & 1e-5 & 0 \\
    & APGCN & 8 & 256 & 0.1 & n\_iter:10; prop\_penalty: 0.05; & 1e-2 & 5 & 1e-5  & 0 \\
    & AutoGCN & 8 & 90 & 0.1 & K:16 & 1e-3 & 5 & 1e-5 & 0
    \\
                    \midrule
    \parbox[t]{2mm}{\multirow{8}{*}{\rotatebox[origin=c]{90}{CIFAR10}}} & MLP & 8 & 162 & 0.1 & - & 1e-3 & 5 & 1e-5 & 0 \\
    & GIN & 8 & 110 & 0.1  & n\_mlp\_GIN:2; learn\_eps\_GIN:True; neighbor\_aggr\_GIN:sum & 1e-3 & 5 & 1e-5 & 0 \\
    & GraphSage & 8 & 110 & 0.1 & sage\_aggregator:mean & 1e-3 & 5 & 1e-5 & 0 \\    
    & GAT & 8  & 19 & 0.1 & n\_heads:8;self\_loop:true & 1e-3 & 5 & 1e-5 & 0  \\
    & MoNet & 8 & 90 & 0.1 & kernel:3; pseudo\_dim\_MoNet:2 & 1e-3 & 5 & 1e-5 & 0 \\
    & ChebNet & 8 & 90 & 0.1 & K:2 & 1e-3 & 5 & 1e-5 & 0\\
    & GCN & 8 & 150 & 0.1 & - & 1e-3 & 5 & 1e-5 & 0 \\
    & SGC & 8 & 150 & 0.1 & - & 1e-2 & 5 & 1e-5 & 0 \\
    & APGCN & 8 & 256 & 0.1 & n\_iter:10; prop\_penalty: 0.05; & 1e-2 & 5 & 1e-5  & 0 \\
    & AutoGCN & 8 & 90 & 0.1 & K:16 & 1e-3 & 5 & 1e-5 & 0
    \\
    
      \bottomrule
    \end{tabular}
    }
  \end{center}
\end{table*}

\theoremstyle{definition}
\begin{definition}[High-pass linear filter] The high-pass linear filter function is defined as
\begin{equation}
F_{high}(\boldsymbol\lambda) = p(a\boldsymbol\lambda+1-2a).
\end{equation}
with $p>0$ and $a\in(0,1)$.

The graph convolutional  kernel of the high-pass filter is then derived as 
\begin{align}
\mathbf{C}_{high} &= \mathbf{U}diag(F_{high}(\boldsymbol\lambda))\mathbf{U}^T\\
& = \mathbf{U}diag(p(a\boldsymbol\lambda+1-2a))\mathbf{U}^T \\
& = p(a\mathbf{L}+\mathbf{I} -2a\mathbf{I} )\\
& = p(-a\mathbf{\tilde{A}}+(1-a)\mathbf{I})
\end{align}

\end{definition}

\theoremstyle{definition}
\begin{definition}[Middle-pass quadractic filter] The middle-pass quadractic filter function is defined as
\begin{equation}
F_{mid}(\boldsymbol\lambda) = p((\boldsymbol\lambda-1)^2-a).
\end{equation}
with $p>0$ and $a\in(0,1]$.

The graph convolutional  kernel of the high-pass filter is then derived as 
\begin{align}
\mathbf{C}_{mid} &= \mathbf{U}diag(F_{mid}(\boldsymbol\lambda))\mathbf{U}^T\\
& = \mathbf{U}diag(p((\boldsymbol\lambda-1)^2-a))\mathbf{U}^T \\
& =  p(\mathbf{U}(diag(\boldsymbol\lambda-1))^2\mathbf{U}^T-a\mathbf{I})\\ 
\end{align}
As $\mathbf{U}^T\mathbf{U}=\mathbf{I}$,
\begin{align}
\mathbf{C}_{mid} &= p(\mathbf{U}diag(\boldsymbol\lambda-1)\mathbf{U}^T\mathbf{U}diag(\boldsymbol\lambda-1)\mathbf{U}^T-a\mathbf{I})\\
& = p((\mathbf{L}-\mathbf{I})^2 -a\mathbf{I} )\\
& = p(\mathbf{\tilde{A}}^2-a\mathbf{I})
\end{align}
\end{definition}

\section{Hyperparameter settings for all experiments.}
The hyper-parameter settings for all experiments are given in Table \ref{tab:hyperparam}. 

\balance

\end{document}